\documentclass[10pt,journal]{IEEEtran}

\usepackage[pdftex]{graphicx}
\usepackage{cite}
\usepackage{amsmath,amsfonts,amsthm}
\usepackage[caption=false,font=footnotesize,labelfont=sf,textfont=sf,subrefformat=parens,labelformat=parens]{subfig}
\usepackage{multirow}
\usepackage{booktabs}
\usepackage[ruled,linesnumbered,procnumbered]{algorithm2e}
\usepackage{enumitem}
\usepackage{etoolbox}
\usepackage{xcolor}
\usepackage{hyperref}

\makeatletter
\AtBeginEnvironment{procedure}{\let\c@algocf\c@procedure}
\makeatother

\hypersetup{
    colorlinks = true,
    urlcolor   = black,
    linkcolor  = black,
    citecolor  = black
}

\begin{document}

\theoremstyle{definition}
\newtheorem{definition}{Definition}

\newcommand{\dnn}{\textbf{MLP}}
\newcommand{\doctorai}{\textbf{DoctorAI}}
\newcommand{\retain}{\textbf{RETAIN}}
\newcommand{\deepr}{\textbf{Deepr}}
\newcommand{\gram}{\textbf{GRAM}}
\newcommand{\mime}{\textbf{MiME}}
\newcommand{\dipole}{\textbf{Dipole}}
\newcommand{\timeline}{\textbf{Timeline}}
\newcommand{\gbert}{\textbf{G-BERT}*}
\newcommand{\hitanet}{\textbf{HiTANet}}
\def\modelname{\textbf{Sherbet}}

\newcommand{\dnnn}{{MLP}}
\newcommand{\doctorain}{{DoctorAI}}
\newcommand{\retainn}{{RETAIN}}
\newcommand{\deeprn}{{Deepr}}
\newcommand{\gramn}{{GRAM}}
\newcommand{\mimen}{{MiME}}
\newcommand{\dipolen}{{Dipole}}
\newcommand{\timelinen}{{Timeline}}
\newcommand{\gbertn}{{G-BERT*}}
\newcommand{\hitanetn}{{HiTANet}}
\def\modelnamen{Sherbet}

\newcommand\norm[1]{\left\lVert#1\right\rVert}

\title{Self-Supervised Graph Learning with Hyperbolic Embedding for Temporal Health Event Prediction}

\author{Chang~Lu,~
        Chandan~K.~Reddy,~\IEEEmembership{Senior~Member,~IEEE,}
        Yue~Ning%
\thanks{C. Lu and Y. Ning are with the Department of Computer Science, Stevens Institute of Technology, New Jersey,
NJ, 07310. E-mail: \href{mailto:clu13@stevens.edu}{clu13@stevens.edu}, \href{mailto:yue.ning@stevens.edu}{yue.ning@stevens.edu}}%
\thanks{C.K. Reddy is with the Department of Computer Science, Virginia Tech, Arlington, VA 22203. E-mail: \href{mailto:reddy@cs.vt.edu}{reddy@cs.vt.edu}}%
}

\maketitle

\begin{abstract}
Electronic Health Records (EHR) have been heavily used in modern healthcare systems for recording patients' admission information to health facilities. Many data-driven approaches employ temporal features in EHR for predicting specific diseases, readmission times, and diagnoses of patients. However, most existing predictive models cannot fully utilize EHR data, due to an inherent lack of labels in supervised training for some temporal events. Moreover, it is hard for existing methods to simultaneously provide generic and personalized interpretability. To address these challenges, we propose \modelname, a self-supervised graph learning framework with hyperbolic embeddings for temporal health event prediction. We first propose a hyperbolic embedding method with information flow to pre-train medical code representations in a hierarchical structure. We incorporate these pre-trained representations into a graph neural network to detect disease complications, and design a multi-level attention method to compute the contributions of particular diseases and admissions, thus enhancing personalized interpretability. We present a new hierarchy-enhanced historical prediction proxy task in our self-supervised learning framework to fully utilize EHR data and exploit medical domain knowledge. We conduct a comprehensive set of experiments on widely used publicly available EHR datasets to verify the effectiveness of our model. Our results demonstrate the proposed model's strengths in both predictive tasks and interpretable abilities.
\end{abstract}


\IEEEpeerreviewmaketitle

\section{Introduction}
\label{sec:intro}
\IEEEPARstart{G}{iven} the promising potential of Electronic Health Records (EHR), mining interpretable predictive patterns from EHR data has significant value in healthcare and has drawn a lot of attention in recent years. EHR data are complex in nature and typically contain sequences of patients' admission records, such as diagnoses, clinical notes, and medications. Effective analysis of EHR data is important for both medical professionals and patients as it can provide preventative health alerts and personalized care plans.

A variety of predictive models using deep learning technology have been proposed for predicting temporal events, such as diagnosis prediction~\cite{choi2016doctor, zheng2020predicting, choi2017gram, bai2018interpretable, xu2021comorbidity}, mortality prediction~\cite{nguyen2018effective, yang2020cross, darabi2019taper, mulyadi2021uncertainty}, risk prediction~\cite{Phuoc2017deepr, huang2017regularized, che2017boosting, wang2019data}, and medication recommendation~\cite{yao2018topic, shang2019pretrain}. A common supervised training approach to utilize EHR data for temporal event prediction is to use previous records as features and the records of next admissions as labels. However, this approach will inherently ignore patients' final admissions due to the lack of labels. Moreover, learning effective representations for medical concepts by effectively leveraging the domain knowledge is still an open problem in healthcare applications. In summary, there are still some challenges for predictive models using temporal information of EHR data.

\begin{figure}
    \subfloat[Supervised training]{
        \includegraphics[width=0.4\linewidth]{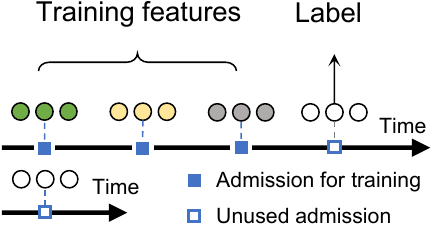}
        \label{fig:label_example_1}
    } \hfill
    \subfloat[Self-supervised training]{
        \includegraphics[width=0.5\linewidth]{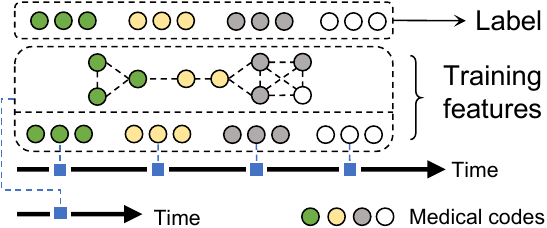}
        \label{fig:label_example_2}
    }
    \caption{{An example of predicting temporal events with supervised training on original EHR data in a temporal setting and with self-supervised training on extracted EHR data in a graph structure.}}
    \label{fig:label_example}
\end{figure}

\noindent \paragraph{Fully utilizing EHR data} A large number of EHR records are unused in traditional supervised training. Final admission records of patients, including single admissions, are discarded for training because the labels are missing for next potential admissions. \figurename~\subref*{fig:label_example_1} shows the reason for the lack of labels while predicting temporal events. For multiple-admission patients (top arrow), the final admission record is used as the label for a prediction and cannot be used as training features. For single-admission patients (bottom arrow), there are no labels because there are no next admissions. However, according to the statistics of a widely used EHR dataset~\cite{johnson2016mimic}, the number of multiple-admission patients is only 20\% of single-admission patients. Hence, a majority of valuable information in EHR data is discarded.

\paragraph{Exploiting disease hierarchies during prediction} In modern disease classification systems like ICD-9-CM~\cite{icd9cm}, diseases are classified into various categories as medical codes in multiple levels and form a hierarchical structure. Existing methods such as GRAM~\cite{choi2017gram} and G-BERT~\cite{shang2019pretrain} mainly use this structure to extract disease features using attention methods. However, this type of domain knowledge is helpful in guiding model predictions. By predicting disease hierarchies, representations of diseases can be further refined.

\paragraph{Learning hidden representations for related diseases} Most existing work treats diseases independently while neglecting disease interactions (i.e., complications). However, such complications are generally crucial in medical practice. For example, longstanding hypertension will eventually lead to heart failure. Consequently, it is very common for patients with heart failure to have been suffering from hypertension prior to being admitted for a heart failure condition~\cite{messerli2017transition}.

\paragraph{Simultaneously providing generic and personalized interpretability} Generic interpretability provides discovered common knowledge such as disease complications from the entire set of patient records. Personalized interpretability refers to explanations for individual patients based on their personal admission records. Both these two kinds of interpretabilities should be considered in healthcare models to make prediction results reliable to doctors and patients. However, current approaches~\cite{choi2017gram, bai2018interpretable, choi2016retain, ma2017dipole} mainly focus on one type of interpretability, while it is critical to simultaneously provide both generic and personalized interpretability.

To address these challenges, we propose \modelname, a \textit{\textbf{s}elf-supervised graph learning framework with \textbf{h}yp\textbf{er}bolic em\textbf{be}ddings for \textbf{t}emporal health event prediction}. As a subset of unsupervised learning methods, the self-supervised learning method in this work is different from other pre-training methods, such as G-BERT~\cite{shang2019pretrain}. We design a special proxy task for self-supervised learning to hierarchically predict historical diagnoses of patients. \figurename~\subref*{fig:label_example_2} shows the manner in which self-supervised learning can fully utilize EHR data. The proxy task constructs an interaction graph  for medical codes in all admissions rather than treating each admission independently. It enables us to incorporate single-admission patients and the final admissions of multiple-admission patients by generating new labels for all admissions. When implementing this task, we first pre-train disease representations using a new hyperbolic embedding method with information flow to reconstruct the disease hierarchical structure. In order to model disease interactions, we next construct a weighted and directed graph for diseases based on their occurrences in patients' admission records. Then we design a graph encoder architecture for self-supervised learning. The first part of the graph encoder is a graph neural network on the constructed graph to extract hidden disease representations and further learn the disease complications. Then, we develop a multi-level attention mechanism as the encoder to learn the representation of admissions and patients from the admission records. The contribution of specific diseases and admissions to a given prediction task can thus be quantified. The self-supervised learning component, which is also the decoder, is designed with the proxy task of hierarchically predicting historical diseases. Finally, we build a fine-tuning module for specific tasks. The main contributions of this work are summarized as follows:
\begin{itemize}
    \item We propose a novel self-supervised graph learning framework and a hierarchy-enhanced historical prediction task to fully exploit the admission records in EHR data and hierarchical structures of medical codes.
    \item We propose a new hyperbolic embedding method with an information flow strategy to pre-train medical code representations using the disease hierarchical structure. It can simultaneously consider hierarchical domain knowledge and similarities among medical codes.
    \item We design a weighted and directed disease interaction graph to learn the disease complications {as generic interpretability}. Together with multi-level attention, the proposed model is able to provide generic and personalized interpretability.
\end{itemize}

The rest of this paper is organized as follows: Section~\ref{sec:related_works} summarizes the related work. Section~\ref{sec:methodology} formally defines the prediction problem and self-supervised learning task. Then, we demonstrate the experimental settings and results in Sections~\ref{sec:experiments} and~\ref{sec:results}. Finally, we summarize our work and discuss potential future research in Section~\ref{sec:conclusion}.
\section{Related Work}
\label{sec:related_works}
\paragraph*{Predictive Models in Healthcare}
Deep learning methods have been widely adopted for learning effective representations of complex and dynamic data in a wide range of applications including temporal event modeling in healthcare. Choi \textit{et al.}~\cite{choi2016doctor} proposed DoctorAI to predict diagnoses in following admissions and the time interval of hospital readmissions using recurrent neural networks (RNN) with gated recurrent unit (GRU~\cite{cho2014learning}) cells. A reverse-time RNN model with attention, RETAIN, was proposed by Choi \textit{et al.} ~\cite{choi2016retain} to predict heart failure and~provide some interpretability of predictive models. Nguyen \textit{et al.}~\cite{Phuoc2017deepr} proposed Deepr which regards diseases as words and admission records as sentences and uses convolutional neural networks (CNN) as a language model to predict re-admission possibilities of patients in the next three months. Ma \textit{et al.}~\cite{ma2017dipole} proposed Dipole using a bi-directional RNN with various attention methods to predict future diagnoses. Bai \textit{et al.}~\cite{bai2018interpretable} considered the time duration between two admissions and proposed the Timeline model. {Luo \textit{et al.}~\cite{luo2020hitanet} used a self-attention-based method to detect key time steps in patients' historical admissions.} However, as discussed above, these predictive models usually do not consider data that lack labels (such as single and last admissions) and thus cannot fully utilize the potential of EHR data.

\paragraph*{Unsupervised and Self-supervised Learning in Healthcare}
Self-supervised learning refers to training models with automatically generated labels~\cite{jing2020self}. 
It is used for obtaining distinguishable features of samples by pre-training the model on proxy tasks. Gidaris \textit{et al.}~\cite{gidaris2018unsupervised} created a pretext task to predict image rotation using ConvNet. Xu \textit{et al.}~\cite{xu2019ternary} proposed a self-supervised framework LabNet by leveraging word vectors of both seen and unseen labels for cross-modal retrieval. A common approach when using EHR data is to treat the diseases and the admission records as words and sentences, respectively. Then popular language models such as Transformer~\cite{vaswani2017attention} and BERT~\cite{devlin2018bert} can be applied to learn the representation of diseases. Choi \textit{et al.}~\cite{choi2017gram} applied GloVe~\cite{pennington2014glove} to initialize disease embeddings in a medical ontology tree using labeled data. Shang \textit{et al.}~\cite{shang2019pretrain} proposed the G-BERT model to recommend medicines for patients considering an admission as a sentence and using BERT to pre-train disease embeddings. However, one problem of applying language models on EHR data is that diagnoses in an admission record typically do not have an ordering (like words in a sentence). Therefore, language models may not fit EHR data well because a different order of diagnoses may lead to significantly different prediction results.

\paragraph*{Graph Neural Networks and Hyperbolic Representations}
{Graph neural networks are developed for data with graph structures. Yu \textit{et al.}~\cite{yu2012adaptive} proposed an adaptive hypergraph learning approach by varying neighborhood size for transductive image classification. Kipf \textit{et al.} proposed graph convolutional networks (GCN)~\cite{kipf2016semi} to generalize convolutional neural networks for node classifications. Nickel \textit{et al.}~\cite{nickel2017poincare} designed a Poincar{\'e} ball model and an optimization method to learn representations for hierarchical data. Chami \textit{et al.}~\cite{chami2020low} leveraged hyperbolic embedding to improve knowledge graph representations in low dimensions. Furthermore, a self-supervised hyperboloid embedding learning method\cite{choudhary2021self}  was proposed to capture hierarchical semantic information in knowledge graphs.} Recently, graph neural networks have also become popular and effective for modeling the EHR data. Choi \textit{et al.}~\cite{choi2017gram} used a medical ontology graph based on hierarchical domain knowledge and applied an attention mechanism to aggregate disease embeddings in different hierarchies. Shang \textit{et al.}~\cite{shang2019pretrain} also utilized this knowledge but designed a two-stage attention method for diseases. Choi \textit{et al.}~\cite{gct_aaai20} proposed GCT, a graph convolutional transformer by constructing a graph of diagnoses, treatments, and lab results. Lu \textit{et al}.~\cite{lu2021collaborative} considered horizontal links in the medical ontology graph and constructed a patient-disease graph to learn hidden disease relations. Most models treat diseases independently, or only apply graph domain knowledge (e.g., hierarchical medical code classification) in feature extractions, but do not take into account a common fact that diseases in different classes can also have strong interactions, i.e., disease complications.

Given these problems in existing work, we develop a method that can utilize more information in EHR data. The proposed method takes disease hierarchical structures and hidden disease relations into account in temporal predictions. In addition, we also focus on providing interpretability from both disease and patient aspects.
\begin{figure}
    \centering
    \includegraphics[width=\linewidth]{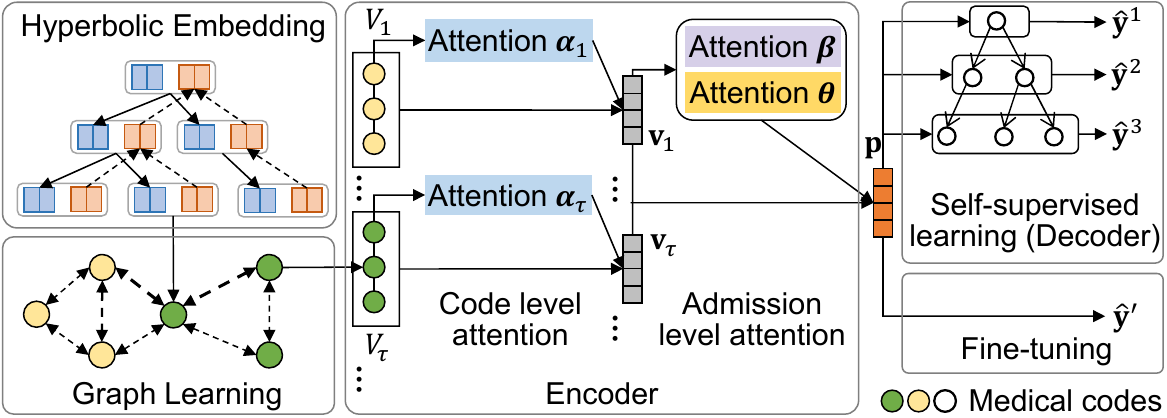}
    \caption{An overview of the proposed {\modelnamen} model. Hyperbolic representations of medical codes are firstly pre-trained with a hierarchical structure. Then, the {\modelnamen} graph is constructed based on the occurrence of medical codes in admission records. A graph neural network is adopted to learn the hidden embedding of medical codes. Next, an encoder including the code level attention and admission level attention is designed to encode the admission records of a patient into a patient vector. For the self-supervised learning, a historical hierarchy prediction task is designed to utilize the hierarchical structure of medical codes in prediction. Finally, after self-supervised learning, we can apply fine-tuning for specific prediction tasks.}
    \label{fig:system}
\end{figure}

\section{The Proposed Methodology}
\label{sec:methodology}

We first describe the basic notations and formulate the problem of predicting temporal events (Section~\ref{sec:method_problem}). Next, we demonstrate an unsupervised learning method to initialize disease representations using hyperbolic embeddings for the disease hierarchical structure (Section~\ref{sec:method:init}). Then we introduce the strategy to construct the disease graph (Section~\ref{sec:method_graph}), followed by the proposed self-supervised learning method using a graph encoder-decoder (Section~\ref{sec:method_pretraining}). The fine-tuning of the model is performed after self-supervised learning (Section~\ref{sec:method_finetuning}). Finally, we discuss the interpretability of \modelnamen~from two perspectives including: disease complications as well as specific contributions of historical diagnoses and admissions to the predictions (Section~\ref{sec:method_interpret}). An overview of the system framework of {\modelnamen} is shown in \figurename{~\ref{fig:system}}.

\subsection{Problem Formulation}
\label{sec:method_problem}

An EHR dataset contains temporal admission records of patients. In each admission, a patient is diagnosed with one or more diseases represented by medical codes, in the format of ICD-9-CM~\cite{icd9cm} or ICD-10~\cite{icd10}. We denote the entire set of medical codes by $\mathcal{C} = \{c_1, c_2, \dots, c_{|\mathcal{C}|}\}$ in the EHR dataset. For a patient $u$, one clinical record $V^u_\tau \subset \mathcal{C}$ is a subset of $\mathcal{C}$, where $\tau = 1, 2, \dots, T^u$ denotes the $\tau$-th admission record of patient $u$ who has a total of $T^u$ admissions. In the rest of this paper, we drop the superscript $u$ in $V^u, T^u$ for better readability unless otherwise stated. The important symbols used in this paper are listed in Table~\ref{tab:symbols}.

\begin{table}
    \centering
    \caption{Notations used in this paper.}
    \label{tab:symbols}
    \begin{tabular}{cl}
        \toprule
        \textbf{Notation} & \textbf{Description}\\
        \midrule
        $\mathcal{D}$ & EHR dataset \\
        $\mathcal{H}, H$ & Disease hierarchical structure and level number \\
        $\mathcal{U}, \mathcal{C}$ & Sets of patients and medical codes\\
        $V_\tau$ & $\tau$-th admission record for a patient \\
        $\mathbf{E}$ & Embeddings of medical codes\\
        $\mathbf{X}$ & Learned hidden embeddings of medical codes\\
        $\mathbf{A}$ & Adjacency matrix of medical codes \\
        $\mathbf{v}_\tau$ & Embedding vector of the $\tau$-th admission \\
        $\mathbf{p}$ & Embedding vector of a patient \\
        \bottomrule
    \end{tabular}
\end{table}

\begin{definition}[\textbf{\textit{EHR dataset}}]
    An EHR dataset is given by $\mathcal{D} = \{ r_u | u \in \mathcal{U} \}$ where $\mathcal{U}$ is the entire set of patients in $\mathcal{D}$, and $r_u = (V_1,V_2, \dots, V_{T})$ is the admission records of patient $u$. Each admission $V_\tau \subset \mathcal{C}$ contains a subset of $\mathcal{C}$.
\end{definition}

\begin{definition}[\textbf{\textit{Health Event Prediction}}]
    \label{def:problem}
    Given an EHR dataset $\mathcal{D}$ and a patient $u$ who has $T$ historical admissions, the goal of the prediction task in this paper is to predict the future health event $\mathbf{y}_{T + 1}$ for patient $u$ such as diagnosis or heart failure prediction.
For instance, if the task is to predict the diagnoses for a patient's $(T + 1)$-th admission, the goal will be estimating the probabilities of all medical codes, i.e.,  $\mathbf{y}_{T+1} \in \{0, 1\}^{|\mathcal{C}|}$ in the $(T + 1)$-th admission of this patient.
\end{definition}

\subsection{Hyperbolic Embedding with Information Flow}
\label{sec:method:init}
The ICD-9-CM system provides a domain knowledge base to classify diseases into various categories represented by medical codes in multiple levels. 
In this system, the medical codes form a hierarchical structure $\mathcal{H}$ with $H$ levels, i.e., a tree. To obtain effective representations of medical codes, we aim to learn the embeddings of medical codes by reconstructing the skeleton of $\mathcal{H}$. We take advantage of the Poincar{\'e} ball model~\cite{nickel2017poincare, dhingra2018embedding} to learn the representations of the hierarchical structure of diseases, which encodes nodes in $\mathcal{H}$ to a hyperbolic space. The distance in the hyperbolic space between embedding vectors $\mathbf{e}_i$ and $\mathbf{e}_j$ of two medical codes $c_i, c_j \in \mathcal{H}$ is defined as:
\begin{align}
    \label{eq:hyperbolic_dist}
    \text{d}(\mathbf{e}_i, \mathbf{e}_j) = \cosh^{-1}\left(1 + 2\frac{\norm{\mathbf{e}_i - \mathbf{e}_j}^2}{(1 - \norm{\mathbf{e}_i}^2(1 - \norm{\mathbf{e}_j}^2)}\right) .
\end{align}
In $\mathcal{H}$, higher level diseases can be regarded as a summary of their children, while lower level diseases provide more precise descriptions for their parents. Following this intuition, we also propose an information flow strategy to model the similarity and distinction among ancestor nodes and children nodes in $\mathcal{H}$. Together with the hyperbolic embedding, we are able to simultaneously consider the hierarchy and similarity of medical codes.

\begin{figure}
    \centering
    \includegraphics[width=0.85\linewidth]{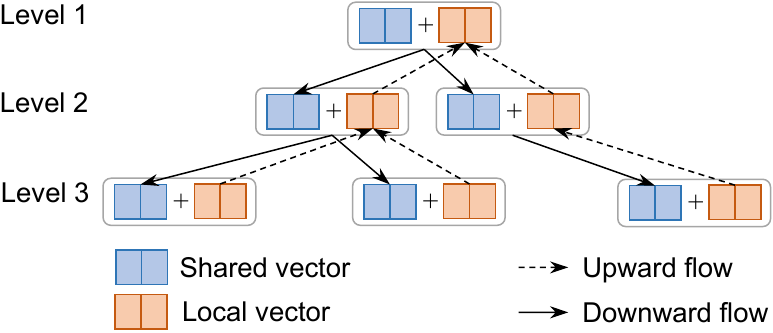}
    \caption{Information flow in disease hierarchical structure. The downward flow starts from a parent node to the shared part of a child node. The upward flow starts from the local part of a child node to the local part of a parent node.}
    \label{fig:hierarchical_embedding}
\end{figure}

In practice, medical codes recorded in EHR datasets are usually leaf nodes. In some cases when a patient is diagnosed with higher-level diseases, i.e., non-leaf nodes, we recursively create virtual child nodes for each non-leaf node to pad them into virtual leaf nodes in the same level. Therefore, in this paper, set $\mathcal{C}$ contains only leaf nodes and virtual leaf nodes in the last level. We use $|\mathcal{H}|$ to denote the total number of medical codes in $\mathcal{H}$. To represent the information flow in $\mathcal{H}$, we first assign each medical code $c_i$ with two randomly initialized trainable embedding vectors, a shared vector $\mathbf{s}_i \in \mathbb{R}^d$ and a local vector $\mathbf{t}_i \in \mathbb{R}^d$. The shared vector $\mathbf{s}_i$ is designed to represent the information inherited from its parent. The local vector $\mathbf{t}_i$ contains private and more precise information of $c_i$ that makes $c_i$ different from its parent. Then, the public embedding vector $\mathbf{e}'_i$ of $c_i$ is calculated as:
\begin{align}
    \label{eq:he}
    \mathbf{e}'_i = \lambda_i \times \mathbf{s}_i + (1 - \lambda_i) \times \mathbf{t}_i  \in \mathbb{R}^d,
\end{align}
where $\lambda_i$ is a trainable coefficient to integrate the shared and local vectors. To capture the information flow in the disease hierarchical structure, we propose a hierarchical embedding method using the shared and local embedding vectors of each medical code. \figurename{~\ref{fig:hierarchical_embedding}} shows the information flow in two directions, i.e., downward and upward flow. A downward flow indicates that the shared vector of a lower level medical code inherit the public vector of its parent, and an upward flow simulates the summary of this medical code's children by aggregating the children's local vectors. Such flows can be summarized as follows:
\begin{alignat}{2}
    \mathbf{s}'_i &= \mathbf{e}'_j &&\quad\text{(Downward flow)}, \\
    \mathbf{t}'_i &= \frac{1}{n_i}\sum_{k = 1}^{n_i}{\mathbf{t}_k} &&\quad\text{(Upward flow)},
\end{alignat}
where $c_j \in \mathcal{H}$ is the parent of $c_i$, $c_k \in \mathcal{H}$ is a child of $c_i$, and $n_i$ is the number of $c_i$'s children. After this flow process, we use Equation~(\ref{eq:he}) on $\mathbf{s}'_i$ and $\mathbf{t}'_i$ to calculate the new public embedding vector $\mathbf{e}_i$.

To reconstruct the hierarchical structure $\mathcal{H}$, we aim to minimize an overall distance among all nodes in $\mathcal{H}$. We calculate the distance between the public embeddings of two medical codes because $\mathbf{e}_i$ contains the information passed from $c_i$'s parent and children. We assume the distance should be small between two connected nodes, while large for unconnected nodes. Let $\mathcal{A} = \{(i, j) \mid c_i, c_j \in \mathcal{H}\}$ be the set of connected node pairs in $\mathcal{H}$, the loss function $\mathcal{L}_{\text{rec}}$ of reconstructing the hierarchical structure using the hyperbolic distance between two node embedding vectors is defined as:
\begin{align}
    \mathcal{L}_{\text{rec}} = - \sum_{(i, j) \in \mathcal{A}}\log{\frac{e^{-\text{d}(\mathbf{e}_i, \mathbf{e}_j)}}{\sum_{j'\ \in \mathcal{N}(i) \cup \{v\}}{e^{-\text{d}(\mathbf{e}_i, \mathbf{e}_{j'})}}}}.
\end{align}
Here, $\mathcal{N}(i) = \{j'\mid(c_i, c_{j'}) \notin \mathcal{A} \}$ denotes the set of non-adjacent nodes for $c_i$. Finally, we use $\mathcal{L}_{\text{rec}}$ to pre-train the representation of medical codes using back-propagation and obtain the final embedding $\mathbf{E} \in \mathbb{R}^{|\mathcal{C}| \times d}$, which is the collection of $\mathbf{e}_i$ of medical codes in the lowest level, i.e., leaf nodes. The pre-trained medical code embeddings $\mathbf{E}$ will be further used in the self-supervised graph learning stage.

The pseudo-code of hyperbolic embedding is shown in Procedure~\ref{algo:hyperbolic_algo}. Lines 2-6 summarize the information flow, and lines 7-9 correspond to the optimization process to pre-train the embeddings using $\mathcal{L}_{\text{rec}}$.
\SetAlFnt{\small}
\begin{procedure}
    \caption{HyperbolicEmbedding($\mathcal{H}$)}
    \label{algo:hyperbolic_algo}
    \DontPrintSemicolon
    \SetAlgoLined
    \SetKwInOut{Input}{Input}
    \SetKwInOut{Output}{Output}

    \Input{The hierarchical structure $\mathcal{H}$ of medical codes}
    \Output{Embeddings $\mathbf{E}$ for leaf nodes of $\mathcal{H}$}
    Randomly initialize shared and local vectors \\
    \For{$c_i \in \mathcal{H}$}{
        $\mathbf{e}'_i \leftarrow$ combine shared and local vector $\mathbf{s}_i, \mathbf{t}_i$ \\
        $\mathbf{s}'_i, \mathbf{t}'_i \leftarrow$ Downward and upward flow \\
        $\mathbf{e}_i \leftarrow$ combine $\mathbf{s}'_i, \mathbf{t}'_i$ \\
    }
    \Repeat{convergence}{
        Calculate distance by Equation~(\ref{eq:hyperbolic_dist}) \\
        Optimizing embedding vectors using $\mathcal{L}_{\text{rec}}$ \\
    }
    \Return $\{\mathbf{e}_i \mid \text{$c_i$ is a leaf node in $\mathcal{H}$}\}$
\end{procedure}

\subsection{{\modelnamen} Graph Construction}
\label{sec:method_graph}
Disease complications are closely related to their co-occurrence frequencies. For example, hypertension and heart failure are often diagnosed on the same patient. Therefore, we design a directed and weighted graph $\mathcal{G}$ to represent the sub-graph patterns of disease complications. In this graph, each vertex $c_i$ of $\mathcal{G}$ is a medical code in $\mathcal{C}$. To describe the complication of two diseases, i.e., medical codes $c_i$ and $c_j$, we create directed edges between vertices $c_i, c_j$ using the following rule: For each medical code pair $(c_i, c_j)$, if $c_i$ and $c_j$  occur once in a record $V_\tau$ of a patient $u$, we add two edges $(i, j)$ and $(j, i)$ to $\mathcal{G}$. Here, we create a directed graph because we think such disease relationships are asymmetric in nature, since two diseases may not have the same influence on each other. One disease $c_i$ could be a major complication of another disease $c_j$, however, the opposite might not be true. Therefore, we use directed and weighted edges to describe such dual relationships between two diseases by quantifying such influence in an adjacency matrix $\mathbf{A} \in \mathbb{R}^{|\mathcal{C}| \times |\mathcal{C}|}$, where each element $\mathbf{A}_{ij}$ is the weight of edge $(i, j)$. To model the dual influence of diseases, we first define a co-occurrence matrix, $\mathbf{B} \in \mathbb{N}^{|\mathcal{C}| \times |\mathcal{C}|}$, initialized with all zeros. When calculating the values of $\mathbf{B}$, we increase the elements $\mathbf{B}_{ij}$ and $\mathbf{B}_{ji}$ by 1 for each co-occurrence pair of $c_i$ and $c_j$ in each admission record of all patients. Let $q_i = \sum_{j = 1}^{|\mathcal{C}|}{\mathbf{B}_{ij}}$ be the sum of the $i$-th row of $\mathbf{B}$, then, the weighted adjacency matrix $\mathbf{A}$ is calculated as follows:
\begin{align}
    \label{eq:adjacent}
    \mathbf{A}_{ij} = \begin{cases}
        \hfil 0 &~\text{if $i = j$ and $q_i \ne 0$}, \\
        \hfil 1 &~\text{if $i = j$ and $q_i = 0$}, \\
        \frac{\mathbf{B}_{ij}}{q_i} &~\text{otherwise.} \\
    \end{cases}
\end{align}
Note that $\mathbf{A}$ is typically not symmetric, which makes $\mathcal{G}$ a weighted and directed graph. The element $\mathbf{A}_{ij}$ measures the frequency of the disease pair $(c_i, c_j)$ in all co-occurrence pairs of $c_i$. A higher frequency implies $c_j$ appears more times along with $c_i$ than other medical codes. Therefore, we can infer $c_j$ has more influence on $c_i$, and $c_j$ is a more common complication of $c_i$ than other diseases.

\subsection{Self-Supervised Graph Learning}
\label{sec:method_pretraining}
\subsubsection{Graph Learning}
Based on the key idea of constructing the disease graph $\mathcal{G}$, the influence of adjacent nodes on each other corresponds to the weights on their edges. Therefore, we adopt a multi-layer graph neural network (GNN) to further learn the hidden representation of medical codes, given GNN's scalability and effective representation power. By incorporating weighted edges, GNN can scale features and pay more attention to the neighbors with higher weights. Given the initial embedding $\mathbf{E}$ of medical codes learned from hyperbolic embedding and the weighted adjacency matrix $\mathbf{A}$, the hidden representation $\mathbf{X}$ of medical codes can be calculated by a multi-layer GNN: $\mathbf{X} = \text{GNN}(\mathbf{A},\mathbf{E})$, {based on graph convolutional networks (GCN)}.

More specifically, we first use the embeddings of all medical codes as inputs: $\mathbf{H}^{(0)} = \mathbf{E}$. Then, the $l$-th GNN layer to aggregate the features of medical codes is described as below:
\begin{align}
    \label{eq:gnn_layer}
    \mathbf{H}^{(l + 1)} = \text{ReLU}\left(\mathbf{\hat{A}}\mathbf{H}^{(l)}\mathbf{W}_g^{(l)} \right).
\end{align}
Here, $\mathbf{H}^{(l)} \in \mathbb{R}^{|\mathcal{C}| \times d^{(l)}}$ and  $\mathbf{H}^{(l + 1)} \in \mathbb{R}^{|\mathcal{C}| \times d^{(l + 1)}}$ are the input and output of the $l$-th layer, respectively. $\mathbf{W}_g^{(l)} \in \mathbb{R}^{d^{(l)} \times d^{(l + 1)}}$ is the weight of the $l$-th layer. In addition, $\mathbf{\hat{A}}$ is a normalized adjacency matrix of $\mathbf{A}$, which is calculated as follows:
\begin{gather*}
    \mathbf{\hat{A}}_{ij} = \frac{\widetilde{\mathbf{A}}_{ij}}{\sum_{j = 0}^{|\mathcal{C}|}{\widetilde{\mathbf{A}}_{ij}}}, \label{eq:adj_matrix} \\
    \widetilde{\mathbf{A}} = (1 - \varphi)\mathbf{A} + \varphi\mathbf{I},
\end{gather*}
where $\mathbf{I}$ is the identity matrix, and $0 < \varphi < 1$ is a hyper-parameter to adjust the weight of self-loops when adding $\mathbf{I}$ to $\mathbf{A}$. Note that, we modify the calculation of $\mathbf{\hat{A}}_{ij}$ and $\widetilde{\mathbf{A}}$ in the original GCN to adjust the importance of center medical codes. A larger $\varphi$ denotes higher importance of a center medical code when aggregating neighbor codes into this medical code.
In the last GNN layer, i.e., the $L$-th layer, we get the output of the graph neural network. We let $\mathbf{X} = \mathbf{H}^{(L)} \in \mathbb{R}^{|\mathcal{C}| \times m}$ be the hidden representation learned by the GNN, where $m = d^{(L)}$ is the dimension of the hidden embedding in the last layer.

\subsubsection{Multi-level attention}
After obtaining the hidden representation $\mathbf{X}$ of all medical codes in $\mathcal{C}$, we want to use all records in the EHR data to further train the model parameters as well as the embedding matrix $\mathbf{E}$ of medical codes. The data includes both single and multiple admission records. For each patient $u$, we use an encoder to encode all admission records of $u$ into a patient vector $\mathbf{p}$ to represent $u$ using $\mathbf{X}$:
 \begin{align}
     \mathbf{p} = \text{Encoder}\left( V_1, V_2, \dots, V_T \mid \mathbf{X} \right) .
 \end{align}
Specifically, we apply a two-level attention mechanism:

\paragraph{Code-level attention} Without loss of generality, we assume an admission record $V_\tau$ contains $n$ medical codes. Then, the embedding $\mathbf{x}_i$ of each medical code $c_i \in V_\tau$ can be looked up with $\mathbf{X}$. We adopt a global attention mechanism~\cite{luong-etal-2015-effective} on the medical codes in $V_\tau$ to aggregate their embeddings to an admission embedding $\mathbf{v}_\tau$ for the $\tau$-th admission:
\begin{align*}
    \mathbf{z}_i &= \text{tanh}\left( \mathbf{W}_c\mathbf{x}_i \right) \in \mathbb{R}^a, \\
    \alpha_\tau^i &= \frac{\exp\left( \mathbf{z}_i^{\top} \mathbf{w}_{\alpha} \right)}{\sum_{j=1}^{n}{\exp\left(\mathbf{z}_j^{\top} \mathbf{w}_{\alpha}\right)}} \in \mathbb{R}, \\
    \mathbf{v}_\tau &= \sum_{i = 1}^{n}{\alpha_\tau^i\mathbf{x}_i} \in \mathbb{R}^{m}.
\end{align*}
Here, $\mathbf{W}_c \in \mathbb{R}^{a \times m}$ is a weight matrix, where $a$ is the attention dimension. $\mathbf{w}_\alpha \in \mathbb{R}^{a}$ is the weight to calculate the attention score $\alpha_\tau^i $ for medical code $c_i$ in an admission. The code attention score $\boldsymbol{\alpha}_\tau = [\alpha_\tau^1, \alpha_\tau^2, \dots, \alpha_\tau^n]$ measures the distribution of medical codes in an admission. Finally, we multiply the score with $\mathbf{x}_i$ and calculate a weighted sum as the admission embedding $\mathbf{v}_\tau$.

\paragraph{Admission-level attention} After calculating the embedding $\mathbf{v}_\tau$ of $\tau$-th admission of a patient, we need to learn the patient embedding using all admissions. First, we project the admission embedding $\mathbf{v}_\tau$ to the patient dimension:
\begin{align}
    \tilde{\mathbf{v}}_\tau = \text{LeakyReLU}\left( \mathbf{W}_u\mathbf{v}_\tau \right) \in \mathbb{R}^p.
\end{align}
Here, $\mathbf{W}_u \in \mathbb{R}^{p \times m}$ is a weight matrix. $p$ is the patient dimension. {We select LeakyReLU~\cite{xu2015empirical} as the activation function based on a hyper-parameter tuning procedure in experiments.}

Similar to code-level attention, we also use the global attention for admissions to aggregate the admission embeddings and calculate the patient embedding $\mathbf{p}$. However, the original global attention can only measure the significance of each admission, while it cannot distinguish the contribution of each admission to the specific dimension of the output. For example, if a model predicts a set of diagnoses in the next admission, given 5 previous admissions, we aim to quantify the contribution of each admission to every predicted code. Therefore, besides the attention score, we also design a coefficient $\boldsymbol{\theta}_\tau$ to quantify such contribution:
\begin{align}
    \mathbf{r}_\tau &= \text{tanh}\left( \mathbf{W}_v\tilde{\mathbf{v}}_\tau \right) \in \mathbb{R}^b, \notag \\
    \beta_\tau &= \frac{\exp\left( \mathbf{r}_\tau^{\top} \mathbf{w}_{\beta} \right)}{\sum_{\tau=1}^{T}{\exp\left(\mathbf{r}_\tau^{\top} \mathbf{w}_{\beta}\right)}} \in \mathbb{R}, \notag \\
    \boldsymbol{\theta}_\tau &= \frac{\exp\left( \mathbf{r}_\tau^{\top} \mathbf{W}_{\theta} \right)}{\sum_{\tau=1}^{T}{\exp\left(\mathbf{r}_\tau^{\top} \mathbf{W}_{\theta}\right)}} \in \mathbb{R}^{p}, \notag \\
    \mathbf{p} &= \sum_{\tau = 1}^{T}{\beta_\tau \boldsymbol{\theta}_\tau \odot \tilde{\mathbf{v}}_\tau} \in \mathbb{R}^{p}. \label{eq:patient_embedding}
\end{align}
Here, $\mathbf{W}_v \in \mathbb{R}^{b \times p}$ is a weight matrix and $\mathbf{w}_{\beta} \in \mathbb{R}^{b}$ is the weight to calculate the attention score $\beta_\tau \in \mathbb{R}$ for $t$-th admission. The attention score $\boldsymbol{\beta} = [\beta_1, \beta_2, \dots, \beta_T]$ measures the distribution within admissions. In addition, $\mathbf{W}_\theta \in \mathbb{R}^{b \times p}$ is the weight to calculate the attention score $\boldsymbol{\theta}_\tau$. The attention score $\boldsymbol{\theta} = [\boldsymbol{\theta}_1, \boldsymbol{\theta}_2, \dots, \boldsymbol{\theta}_T]$ is the distribution of $\tilde{\mathbf{v}}_\tau$ over each dimension of the patient embedding. Finally, we calculate a weighted sum of $\tilde{\mathbf{v}}_\tau$ as the patient embedding $\mathbf{p}$. Specifically, the weight is calculated by the multiplying $\beta_\tau$ and $\boldsymbol{\theta}_\tau$, and $\odot$ is the element-wise multiplication. We propose Equation~(\ref{eq:patient_embedding}) because we want to simultaneously measure the importance ($\beta_\tau$) of an admission compared to other admissions, and the contribution ($\boldsymbol{\theta}_\tau$) of this admission to the output. We will further elaborate this idea in Section~\ref{sec:method_interpret}, {Model Interpretability}.

Procedure~\ref{algo:encoder_algo} summarizes the pseudo-code of the multi-level attention encoder. 
Lines 2-4 calculate admission vectors using the code-level attention. Then the patient vector $\mathbf{p}$ is computed using the admission level attention at line 5.
\begin{procedure}

    \caption{Encoder($u, \mathcal{D}, \mathbf{X}$)}
    \label{algo:encoder_algo}
    \DontPrintSemicolon
    \SetAlgoLined
    \SetKwInOut{Input}{Input}
    \SetKwInOut{Output}{Output}

    \Input{A patient $u$, an EHR dataset $\mathcal{D}$, the GNN output of medical code embeddings $\mathbf{X}$}
    \Output{The patient embedding vector $\mathbf{p}$ of $u$}
    $V_1, V_2, \dots, V_{T} \leftarrow$ Get admission records of $u$ from $\mathcal{D}$ \\
    \For{$\tau \leftarrow 1$ to $T$}{
        $\mathbf{v}_\tau \leftarrow$ Calculate the admission embedding using the code-level attention with $\mathbf{X}$ \\
    }
    $\mathbf{p} \leftarrow$ Calculate the patient embedding using the admission-level attention \\
    \Return $\mathbf{p}$
\end{procedure}

\subsubsection{Self-supervised learning}
To leverage all records in~an EHR dataset, we need to utilize records of single-admission patients and final admission records of multiple-admission patients. Since these records lack labels regarding their next potential admissions, we focus on reconstructing historical diagnoses using the patient embedding $\mathbf{p}$. We conjecture that the learned patient representation will be able to reflect the historical admission records of this patient. Therefore, recovering the historical diagnoses of this patient takes advantage of the complete dataset and further optimizes the hidden representation of medical codes.

Assuming the historical diagnoses set is denoted as $\mathcal{V}$, we aim to decode the medical codes from $\mathbf{p}$ into the probability distribution $\hat{\mathbf{y}}$: $\hat{\mathbf{y}}_i = P(c_i \in \mathcal{V} \mid \mathbf{p})$. A na\"ive method is using a multilayer perceptron (MLP) to simulate the function: $\hat{\mathbf{y}} = \sigma\left(MLP\left(\mathbf{p}\right)\right)$. This method directly predicts the distribution from $\mathbf{p}$ but does not utilize the hierarchical structure of medical codes. We can make level-wise predictions by calculating the conditional probability of lower levels, once we get the probabilities of higher levels. Therefore, we design a \textit{\textbf{hierarchy-enhanced historical prediction}} task to incorporate the hierarchical structure of medical codes.

\begin{definition}[\textbf{\textit{Historical hierarchy of diagnoses}}]
    Given a patient $u$, the historical hierarchy of diagnoses in the $h$-th level of $\mathcal{H}$ is defined as $\mathcal{V}^h = {\{\rho^h_{c_i} \mid c_i \in \bigcup_{\tau = 1}^{T}V_\tau\}}$, where $\rho^{h}_{c_i}$ denotes the ancestor of $c_i$ in $h$-th level. At the leaf level, we have $\rho^{H}_{c_i} = c_i$.
\end{definition}
Here, $\mathcal{V}^h$ is the set of medical codes in level $h$ that a patient has ever been diagnosed during their previous $T$ admissions.

\begin{definition}[\textbf{\textit{Hierarchy-enhanced historical prediction}}]
    Given a patient $u$, this task is to predict the probability distribution $\hat{\mathbf{y}}^h \in \mathbb{R}^{n_h}$ of ground-truth labels in the $h$-th level $(h=1,...,H)$, where $n_h$ is the number of medical codes in $h$-th level, and $n_H = |\mathcal{C}|$.
\end{definition}

This is a multi-label classification task for each level. Given a code $c_i  = \rho^h_{c_i}$ in the $h$-th level, according to the directed graphical model for joint probability~\cite{bishop2006pattern}, the joint probability $\hat{\mathbf{y}}_j^h = P(\rho^h_{c_i} \in \mathcal{V}^h, \rho^{h - 1}_{c_i} \in \mathcal{V}^{h - 1}, \dots, \rho^{1}_{c_i} \in \mathcal{V}^{1} \mid \mathbf{p})$ can be calculated as follows:
\begin{align}
    \hat{\mathbf{y}}_{j}^h =&~P(\rho^h_{c_i} \in \mathcal{V}^h, \rho^{h - 1}_{c_i} \in \mathcal{V}^{h - 1}, \dots, \rho^{1}_{c_i} \in \mathcal{V}^{1} \mid \mathbf{p}) \notag \\
    =&~P(\rho^1_{c_i} \in \mathcal{V}^1 \mid \mathbf{p}) 
    \prod_{k = 2}^{h}{P(\rho^{k}_{c_i} \in \mathcal{V}^k \mid \rho^{k - 1}_{c_i} \in \mathcal{V}^{k - 1}, \mathbf{p})} \notag \\
    =&~\prod_{k = 2}^{h}{P(\rho^{k}_{c_i} \in \mathcal{V}^k \mid \rho^{k - 1}_{c_i} \in \mathcal{V}^{k - 1}, \mathbf{p})}. \notag
\end{align}
Here, $P(\rho^1_{c_i} \in \mathcal{V}^1 \mid \mathbf{p}) = 1$ because $\mathcal{V}^1$ only contains the root of $\mathcal{H}$. Then, we use a dense layer to calculate  each conditional probability:
\begin{align}
    P(\rho^{k}_{c_i} \in \mathcal{V}^k \mid \rho^{k - 1}_{c_i} \in \mathcal{V}^{k - 1}, \mathbf{p}) = \sigma(\mathbf{w}_{k}\mathbf{p})_{\rho^{k}_{c_i} } \notag
\end{align}
where $\mathbf{w}_k \in \mathbb{R}^{n_{k} \times p}$ decodes the patient embedding $\mathbf{p}$ to the probability of the medical codes in the $k$-th level level, and $n_k$ is the number of medical codes in $k$-th level.

Finally, the decoder with the hierarchical prediction are defined as follows:
\begin{align}
    \label{eq:decoder}
    \hat{\mathbf{y}}_j^h &= \text{Decoder}(\mathcal{H}, \mathbf{p}) = \prod_{k = 2}^{h}{\sigma(\mathbf{w}_{k}\mathbf{p})}_{\rho^{k}_{c_i} },
\end{align}
\begin{align}
    \label{eq:loss_pretrain}
    \mathcal{L} = \frac{1}{H - 1}\sum_{h = 2}^{H}\frac{1}{n_h}{\text{CrossEntropy}\left(\hat{\mathbf{y}}^h, \mathbf{y}^h\right)}.
\end{align}
Here, $\mathbf{y}^h$ is the ground-truth of ancestor medical codes in the $h$-th level. $\mathbf{y}^h_j = 1$ means $c_j$ is the ancestor $\rho^h_{c_i}$ of $c_i$ in the $h$-th level and $c_j \in \mathcal{V}^h$. Note that, as a self-supervised learning problem, this proxy task is not a simple input reconstruction. It predicts the hierarchy of a set of diagnoses given a sequence of admissions. Hence, it is different from G-BERT~\cite{shang2019pretrain}, which is a pre-training method and reconstructs the diagnoses and medicines given the same input.

\subsection{Fine-tuning and Inference}
\label{sec:method_finetuning}
After self-supervised learning, we obtain the learned embedding vectors of medical codes and model parameters. For a specific task given the same format of input, we first calculate the patient embedding $\mathbf{p}$. Then, we use a fully-connected layer for the real prediction task in Definition~\ref{def:problem}. Finally, we calculate the estimated output $\hat{\mathbf{y}}'$ and the fine-tuning loss $ \mathcal{L}'$ to optimize the encoder model including the medical code embeddings and model parameters:
\begin{align}
    \hat{\mathbf{y}}' &= g\left(\mathbf{W}\mathbf{p}\right) \label{eq:prediction} \in \mathbb{R}^o, \\
    \mathcal{L}' &= \text{Loss}\left(\mathbf{y}', \hat{\mathbf{y}}', \Theta\right) ,\label{eq:loss_train}
\end{align}
where $\mathbf{W} \in \mathbb{R}^{o \times p}$ is the weight matrix for the output and $o$ represents the output size.  $g$ denotes the activation function. $\mathbf{y}'$ is the ground-truth label and $\Theta$ is the set of parameters of the model. Note that $\mathbf{y}', g, o$, and the loss function depend on specific tasks. When optimizing the model with back-propagation, we still keep the embedding matrix and parameters learnable.

The pseudo-code of {\modelnamen} including self-supervised learning and fine-tuning is summarized in Algorithm~\ref{algo:model_algo}. It first pre-trains the hyperbolic embeddings of medical codes at line 1. Then the {\modelnamen}-Graph is constructed and a graph neural network is applied at lines 2-8. The self-supervised learning framework calculates the patient vector for each patient in $\mathcal{D}$, predicts the historical hierarchy of diagnoses, and optimize the model at lines 9-17. Finally, fine-tuning is performed on specific tasks in lines 18-23.

\begin{algorithm}
    \caption{\modelnamen-Optimization($\mathcal{H}, \mathcal{D}$)}
    \label{algo:model_algo}
    \DontPrintSemicolon
    \SetAlgoLined
    \newcommand\mycommfont[1]{\footnotesize\ttfamily\textcolor{gray}{#1}}
    \SetCommentSty{mycommfont}
    \SetKwInOut{Input}{Input}
    \SetKwInOut{Output}{Output}
    \SetKwProg{Proc}{Procedure}{}{}
    \SetKwFunction{HyperbolicEmbedding}{HyperbolicEmbedding}
    \SetKwFunction{Encoder}{Encoder}
    \Input{The hierarchical structure $\mathcal{H}$ of medical codes, an EHR dataset $\mathcal{D}$}
    $\mathbf{E} \leftarrow$ \HyperbolicEmbedding{$\mathcal{H}$} \\
    $\mathbf{\hat{A}} \leftarrow$ Construct graph $\mathcal{G}$ and calculate normalized adjacency matrix by Equation~(\ref{eq:adj_matrix}) \\
    $L \leftarrow$ Graph layer number \\
    $\mathbf{H}^{(0)} \leftarrow \mathbf{E}$ \\
    \For(\tcp*[f]{Graph learning}){$l \leftarrow 0$ to $L - 1$}{
        $\mathbf{H}^{(l + 1)} \leftarrow$ Aggregate $\mathbf{H}^{(l)}$ by Equation~(\ref{eq:gnn_layer})
    }
    $\mathbf{X} \leftarrow \mathbf{H}^{(L)}$ \\
    \Repeat(\tcp*[f]{Self-supervised learning}){convergence}{
        $u \leftarrow$ a patient in $\mathcal{D}$ \\
        $\mathbf{p} \leftarrow$ \Encoder{$u, \mathcal{D}, \mathbf{X}$} \\
        $H \leftarrow$ level number of $\mathcal{H}$ \\
        \For{$h \leftarrow 2$ to $H$}{
            $\mathbf{\hat{y}}^h \leftarrow$ Predict level $h$'s labels by Equation~(\ref{eq:decoder}) \\
        }
        Optimizing the model using $\mathcal{L}$ \\
    }
    \Repeat(\tcp*[f]{Fine-tuning}){convergence}{
        $u \leftarrow$ a patient in $\mathcal{D}$ with multiple admissions \\
        $\mathbf{p} \leftarrow$ \Encoder{$u, \mathcal{D}, \mathbf{X}$} \\
        $\mathbf{\hat{y}}' \leftarrow$ Predict labels based on specific tasks \\
        Optimizing the model using $\mathcal{L}'$ \\
    }
\end{algorithm}

\subsection{Model Interpretability}
\label{sec:method_interpret}
Another advantage of \modelnamen~is its ability to provide interpretable prediction results and learned representations (which is an important part in healthcare). Some existing models~\cite{choi2017gram, bai2018interpretable} only focus on learning medical concept representations while neglecting personalized interpretability at the patient level. On the other hand, a few other models~\cite{ma2017dipole} are able to interpret at the patient level while lacking generic interpretability for common medical knowledge. In \modelnamen, we provide generic interpretability by learning effective disease~representations as well as personalized interpretability by quantifying contributions of each disease and admission for a patient.

\paragraph*{Representation for medical codes} We use the output $\mathbf{X}$ of GNN as the hidden representation of medical codes, since the proposed self-supervised graph learning framework helps to learn effective disease relations {and captures reasonable disease complications as generic interpretability}. We demonstrate the results of the learned representations in Section~\ref{sec:exp_interpret}.

\paragraph*{Code-level contribution} We use the code-level attention scores $\boldsymbol{\alpha}_\tau = [\alpha_\tau^1, \alpha_\tau^2, \dots, \alpha_\tau^n]$ to represent the contributions of medical codes in an admission, since $\mathbf{\alpha}$ is a probability distribution of medical codes.

\paragraph*{Admission-level contribution} Similar to the code-level contribution, we use $\boldsymbol{\beta}$ as a part of the admission-level contribution. Furthermore, we want to quantify the contribution of a specific admission to the output. For example, if the task is to predict all diagnoses in the next admission, we want to comprehend the contribution of the $t$-th admission to each predicted medical code, i.e., $\hat{\mathbf{y}}_i$. Therefore, we define a coefficient $\boldsymbol{\delta} = [\boldsymbol{\delta}_1, \boldsymbol{\delta}_2, \dots, \boldsymbol{\delta}_T]$ to measure this contribution:
\begin{align}
    \boldsymbol\delta_\tau = \beta_\tau\frac{\exp\left(\mathbf{W}\boldsymbol{\theta}_\tau\right)}{\sum_{\tau = 1}^{T}{ \exp\left(\mathbf{W}\boldsymbol{\theta}_\tau\right)}} \in \mathbb{R}^o,
    \label{eq:coeff}
\end{align}
where $\mathbf{W}$ is the weight in Equation~(\ref{eq:prediction}). In this equation, firstly, $\mathbf{W}\boldsymbol{\theta}_\tau$ projects the $\tau$-th admission weight into the output dimension. Then it is multiplied by $\beta_\tau$, so that we can measure the contribution from different admissions. Intuitively, $\boldsymbol\delta_\tau$ first calculates the contribution of each admission ($\beta_\tau$), then it assigns this contribution to each dimension of the output ($\mathbf{W}\boldsymbol{\theta}_\tau$). Finally, the code-level and admission-level contributions can provide personalized interpretability for different patients.
\section{Experimental Setup}
\label{sec:experiments}
\subsection{Tasks and Evaluation Metrics}
We conduct our experiments on two tasks following the settings of GRAM~\cite{choi2017gram}:
\begin{itemize}
    \item \textit{Diagnosis prediction.} This task predicts all medical codes in the next admission. It is a multi-label classification.
    \item \textit{Heart failure prediction.} This task predicts whether patients will be diagnosed with heart failure (HF) in the next admission. It is a binary classification.
\end{itemize}

The evaluation metrics for diagnosis prediction are weighted $F_1$ score (w-$F_1$) as in Timeline~\cite{bai2018interpretable} and top-$k$ recall (R@$k$) as in DoctorAI~\cite{choi2016doctor}. w-$F_1$ is a weighted sum of the $F_1$ score for each class, which measures an overall prediction performance on all classes. R@$k$ is the ratio of true positive numbers in top-$k$ predictions to the total number of positive samples, which measures the prediction performance on a subset of classes.

The metrics to evaluate the HF prediction are the area under the ROC curve (AUC) and $F_1$ score since the HF prediction is a binary classification on imbalanced test data. 

\subsection{Datasets}
We use the MIMIC-III dataset~\cite{johnson2016mimic} in both tasks and the eICU dataset \cite{pollard2018eicu} in the first task to evaluate the performance of our model. MIMIC-III contains 58,976 de-identified admission records between 2001 and 2012 from 46,520 patients. In each record, the diseases are encoded by the ICD-9-CM system. There are 6,981 medical codes in both single and multiple admissions. The eICU dataset records the patients' admissions to intensive care units (ICU). For eICU, a patient may have multiple visits to hospitals, and in each hospital visit, there may be multiple admissions to ICU. However, there are not timestamps for hospital visits in the eICU dataset. Therefore, we regard one hospital visit as an independent patient, and each ICU admission as admission records $V$. In addition, we remove diseases in eICU that cannot be found in the ICD-9-CM or ICD-10 system. Table~\ref{tab:dataset} shows the basic statistics of single and multiple admissions in the MIMIC-III and eICU datasets.

\begin{table}
    \centering
    \caption{Statistics of MIMIC-III and eICU datasets for both single (S) and multiple (M) admissions (adm.).} 
    \begin{tabular}{p{2.7cm}cccc}
        \toprule
        \multirow{2}{*}{\textbf{Dataset}} & \multicolumn{2}{c}{\textbf{MIMIC-III}} & \multicolumn{2}{c}{\textbf{eICU}} \\ \cmidrule{2-5}
         & \textbf{S} & \textbf{M} & \textbf{S} & \textbf{M} \\
        \midrule
        \# patients             & 38,980   & 7,493  & 117,752 & 9,408 \\
        \# adm.                   & 38,980   & 19,894 & 117,752 & 20,209 \\
        Max. \# adm.             & 1        & 42     & 1       & 7 \\
        Avg. \# adm.             & 1        & 2.66   & 1       & 2.15 \\
        \midrule
        \# codes                & 6,427    & 4,880  & 837     & 686 \\
        Max \# codes per adm.      & 39       & 39     & 58      & 54 \\
        Avg \# codes per adm.      & 10       & 13     & 3.47    & 4.40 \\
        \bottomrule
    \end{tabular}
    \label{tab:dataset}
\end{table}

We randomly split the EHR data with multiple admissions into training/validation/test data, which contain 6000/493/1000 patients for MIMIC-III, and 8000/408/1000 patients for eICU. The {\modelnamen} graph and self-supervised learning are constructed and trained with single admission data and training data with multiple admissions to guarantee there is no leakage of test data in fine-tuning. We use all medical codes for self-supervised learning on both tasks. For the diagnosis prediction task, we predict medical codes appearing in multiple admissions. For the HF prediction task, there are 38.5\% positive samples and 61.5\% negative samples in MIMIC-III.

\begin{table*}
    \caption{Diagnosis prediction results on MIMIC-III and eICU using w-${F_1}$ (\%) and R@${k}$ (\%).}
    \label{tab:result_code}
    \centering
    \begin{tabular}{lcccc|cccc}
        \toprule
        \multirow{2}{*}{\textbf{Models}} & \multicolumn{4}{c}{\textbf{MIMIC-III}} & \multicolumn{4}{c}{\textbf{eICU}} \\ \cmidrule{2-9}
        & \multirow{1}{*}{\textbf{w-}$\boldsymbol{F_1}$} & \textbf{R@10}& \textbf{R@20}& \multirow{1}{*}{ \textbf{\# Params}} & \multirow{1}{*}{\textbf{w-}$\boldsymbol{F_1}$} & \textbf{R@10}& \textbf{R@20} & \multirow{1}{*}{ \textbf{\# Params}} \\ 
        \midrule
        \dnn      & 11.68 (0.12) & 26.01 (0.04) & 26.99 (0.03) & 1.38M & 39.45 (0.08) & 63.52 (0.12) & 71.59 (0.09) & 0.40M \\
        \doctorai & 12.04 (0.04) & 25.69 (0.11) & 27.21 (0.08) & 2.53M & 64.13 (0.28) & 77.08 (0.32) & 81.79 (0.26) & 0.50M \\
        \retain   & 18.37 (0.79) & 32.12 (0.82) & 32.54 (0.62) & 2.90M & \textbf{65.04 (1.03)} & 78.68 (0.92) & 83.48 (1.08) & 0.85M \\
        \deepr    & 11.68 (0.02) & 26.47 (0.02) & 27.53 (0.11) & 1.16M & 45.89 (0.21) & 67.63 (0.15) & 74.01 (0.18) & 0.19M \\
        \gram     & 20.78 (0.14) & 34.17 (0.19) & 35.46 (0.26) & 1.59M & 57.95 (0.05) & 75.67 (0.03) & 81.52 (0.03) & 0.41M \\
        \dipole   & 14.66 (0.20) & 28.73 (0.20) & 29.44 (0.20) & 2.18M & 58.41 (0.18) & 75.66 (0.24) & 81.06 (0.19) & 0.56M \\
        \timeline & 16.04 (0.72) & 30.73 (0.79) & 32.34 (0.90) & 1.23M & 57.04 (1.23) & 74.40 (1.17) & 79.86 (0.86) & 0.30M \\
        \gbert    & 22.28 (0.25) & 35.62 (0.18) & 36.46 (0.15) & 5.63M & 63.61 (0.33) & 77.63 (0.44) & 80.06 (0.39) & 2.62M \\
        {\hitanet}  & 21.15 (0.19) & 34.68 (0.25) & 35.97 (0.13) & 3.33M & 62.83 (0.38) & 77.02 (0.41) & 81.37 (0.46) & 1.18M  \\
        \midrule
        \modelname & {\textbf{25.74 (0.04)}} & {\textbf{40.46 (0.08)}} & {\textbf{41.08 (0.08)}} & 1.23M & {64.18 (0.08)} & {\textbf{80.05 (0.09)}} & {\textbf{84.43 (0.11)}} & 0.17M \\
        \bottomrule
    \end{tabular}
\end{table*}

\subsection{Comparison Methods}
We select the following state-of-the-art models to compare the performance with \modelnamen:\footnote{We do not compare with MiME~\cite{choi2018mime}, GCT~\cite{gct_aaai20}, and MPVAA~\cite{chowdhury2019mixed} since we do not use treatment, lab result, and demographic features in this work.}
\begin{itemize}
    \item \dnn:~A deep neural networks consists of 3 layers, and uses multi-hot vectors for medical codes in an admission.
    \item \doctorai~\cite{choi2016doctor}: An RNN-GRU model, which also uses multi-hot vectors as inputs.
    \item \retain~\cite{choi2016retain}: A network of two RNNs with reverse time and attention methods. The inputs are multi-hot vectors for medical codes.
    \item \deepr~\cite{Phuoc2017deepr}: A CNN model which uses the embedding of medical codes as inputs.
    \item \gram~\cite{choi2017gram}: An RNN model with a medical ontology graph. The inputs are medical code embeddings.
    \item \dipole~\cite{ma2017dipole}: A bi-directional RNN model with attention. The inputs are multi-hot vectors.
    \item \timeline~\cite{bai2018interpretable}: An RNN model with attention, using the time duration information. The inputs are the embeddings of medical codes.
    \item \gbert~\cite{shang2019pretrain}: A BERT-based model with a medical ontology graph. It first pre-trains the model on a self-prediction and dual-prediction task, then fine-tunes the model to predict the medicine. We modify \gbertn~by removing the medication module and changing the fine-tuning module to predict diagnoses and heart failure.
    \item {\hitanet~\cite{luo2020hitanet}: A Transformer-based model considering time intervals between admissions. The inputs are multi-hot vectors.}
\end{itemize}

\subsection{Parameter Settings for \modelnamen}
We randomly initialize all embeddings and model parameters. The embedding sizes $d, m, p$ for $\mathbf{E}, \mathbf{X}$, $\mathbf{p}$ are 128, 64, and 64, respectively. We use one graph layer where the hidden unit number is $64$. The attention sizes $a, b$ for code-level and admission-level attention are 64 and 32, respectively. The weight $\varphi$ on the adjacency matrix is 0.9. In addition, for fine-tuning, we add Dropout~\cite{srivastava2014dropout} on the graph layer and the input of the decoder. The graph layer's dropout rates of two tasks on MIMIC-III and diagnosis prediction on eICU are 0.2, 0.8, and 0.2, respectively. The decoder's dropout rates are 0.02, 0.15, and 0.15, respectively. For the fine-tuning of both tasks, the activation function is sigmoid; and loss function is cross-entropy loss.

We use 500 epochs for pre-training hyperbolic embeddings. For self-supervised learning, we use 1000 and 300 epochs on the MIMIC-III dataset and the eICU dataset, respectively, and 200 epochs for fine-tuning with a learning rate decay strategy. The initial learning rate for hyperbolic embedding is 0.01 and decays by 0.01 every 100 epochs. The initial learning rate for self-supervised learning is 0.01 and decays at the (100, 500)-th epoch by 0.1 on the MIMIC-III dataset, and decays at the (100, 250)-th epoch by 0.1 on the eICU dataset. The initial learning rate for fine-tuning is 0.01 and decays at (20, 35, 100)-th/(50, 60, 100)-th epoch for diagnosis prediction on MIMIC-III/eICU, and decays at (25, 40, 45) epochs for HF prediction on the MIMIC-III dataset. We use the Adam optimizer~\cite{KingmaB14} for hyperbolic embedding and RMSProp optimizer~\cite{hinton2012neural} for self-supervised learning and fine-tuning. The batch sizes for hyperbolic embedding, self-supervised learning, and fine-tuning are 256, 128, and 32, respectively.

All programs are implemented using Python 3.7.4 and~Tensorflow 2.3.0 with CUDA 10.1 on a machine with Intel i9-9900K CPU, 64GB memory, and Geforce RTX 2080 Ti GPU.\footnote{The source code of the {\modelnamen} model can be found at \url{https://github.com/LuChang-CS/sherbet}.}

\subsection{Parameter Settings for Baselines}
\begin{itemize}
    \item \dnn: The hidden units for two hidden layers are 128 and 64, respectively.
    \item \doctorai: The hidden size for RNN is 128.
    \item \retain: The embedding size for admissions is 256. The hidden layer size for two RNN layers is 128.
    \item \deepr: The embedding size for medical codes is 100. The kernel size and filter number for an 1-D CNN layer are 3 and 128, respectively.
    \item \gram: The embedding size for medical codes and attention size are 100. The hidden size for RNN is 128.
    \item \dipole: The embedding size for admissions is 256. The concatenation-based attention size is 128. The hidden size for RNN is 128.
    \item \timeline: The embedding size for medical codes and attention size are 100. The hidden size for RNN is 128.
    \item \gbert: The parameter settings are the same as given in  \cite{shang2019pretrain}. (1) GAT part: The input embedding size is 75, the number of attention heads is 4; (2) BERT part: the hidden size is 300. The position-wise feed-forward networks include 2 hidden layers with 4 attention heads for each layer. The dimension of each hidden layer is 300.
    \item {\hitanet: The parameter settings are the same as \cite{luo2020hitanet}. (1) {\hitanetn} part: the dense space size for diseases is 256. The space size for time interval, query vector size, and latent space size for time interval are 64. (2) Transformer part: the dimension of attention embedding is 64. The multi-head number is 4. The size of middle feed-forward network is 1024.}
\end{itemize}
\section{Experimental Results}
\label{sec:results}

\subsection{Prediction Results}

Table~\ref{tab:result_code} shows the results of the weighted $F_1$ score and top $k$ recall on the diagnosis prediction task using both datasets. We select $k = [10, 20]$ to calculate the R@$k$, because the average medical code numbers of an admission are 13 and 4.40 in MIMIC-III and eICU, respectively. In Table~\ref{tab:result_code}, we can observe that \modelnamen~outperforms the baseline models in most cases. Note that, the results of models on eICU are much better than MIMIC-III. This is primarily due to the fact that eICU has only 686 medical codes, while MIMIC-III has 4,880. The comparison among {\modelnamen}, {\gbertn}~and \gramn~indicates that self-supervised learning significantly improves the prediction results on both datasets, since they all use medical ontology information.

\begin{table}
    \caption{HF prediction results on MIMIC-III using AUC (\%) and ${F_1}$ (\%).}
    \label{tab:result_hf}
    \centering
    \begin{tabular}{lccc}
        \toprule
        \textbf{Models} & \textbf{AUC} & $\boldsymbol{F_1}$ & \textbf{\# Params} \\
        \midrule
        \dnn      & 80.09 (0.06) & 68.17 (0.05) & 0.64M \\
        \doctorai & 82.25 (0.02) & 68.42 (0.02) & 1.91M \\
        \retain   & 82.73 (0.21) & 71.12 (0.37) & 1.67M \\
        \deepr    & 81.29 (0.01) & 68.42 (0.01) & 0.53M \\
        \gram     & 82.82 (0.06) & 71.43 (0.05) & 0.96M \\
        \dipole   & 81.66 (0.07) & 70.01 (0.04) & 0.92M \\
        \timeline & 80.75 (0.46) & 69.81 (0.34) & 0.95M \\
        \gbert    & 83.61 (0.18) & 72.37 (0.46) & 2.69M \\
        {\hitanet} & 82.77 (0.35) & 71.93 (0.29) & 2.08M \\
        \midrule
        \modelname & \textbf{86.04 (0.16)} & \textbf{74.27 (0.07)} & 0.91M \\
        \bottomrule
    \end{tabular}
\end{table}

Table~\ref{tab:result_hf} demonstrates the AUC and $F_1$ scores of the results for the HF prediction task. We use only MIMIC-III because some diseases are removed in the eICU dataset. Similar to the diagnosis prediction task, {\modelnamen} obtains the superior performance in terms of AUC and $F_1$. In this task, compared to the best baseline model, the improvement of {\modelnamen} is not as significant as the diagnosis prediction. We infer that the self-supervised learning is able to learn a general knowledge representation of diseases, while HF is only one of the many diseases. On the contrary, as a multi-label classification task, the diagnosis prediction task can fully utilize the representation learned by the self-supervised learning, and distinguish each type of diseases in the output.

In summary, the prediction results of diagnoses and heart failure demonstrate the superior performance of the proposed framework over state-of-the-art models.

\subsection{Ablation Study}
To study the effectiveness of each module in \modelnamen, we also compare five types of \modelnamen~with some modules removed or replaced:
\begin{itemize}
    \item \modelname${_b}$: {\modelnamen} with self-supervised learning and hierarchical prediction, removing hyperbolic embedding.
    \item \modelname${_c}$: {\modelnamen} with self-supervised learning, removing hyperbolic embedding and hierarchical prediction.
    \item \modelname${_d}$: {\modelnamen} with hyperbolic embedding, removing self-supervised learning, {while only using fine-tuning with multiple admission data}.
    \item \modelname${_e}$: {\modelnamen} removing hyperbolic embedding and self-supervised learning, {while only using fine-tuning with multiple admission data}.
    \item \modelname${_f}$: {\modelnamen} with the multi-level attention encoder replaced by T-LSTM~\cite{baytas2017patient}.
    \item {\modelname${_g}$: {\modelnamen} removing the graph structure and GNN by directly using the learned hyperbolic embedding for multi-level attention.}
\end{itemize}
\begin{table}
    \caption{Prediction results of the variants of {\modelnamen}.}
    \label{tab:ablation}
    \centering
    \begin{tabular}{lcccc|cc}
        \toprule
        \multirow{3}{*}{\textbf{Model}} & \multicolumn{4}{c}{\textbf{Diagnosis}} & \multicolumn{2}{c}{\multirow{2}{*}{\textbf{HF}}} \\ \cmidrule{2-5}
        & \multicolumn{2}{c}{\textbf{MIMIC-III}} & \multicolumn{2}{c}{\textbf{eICU}} & & \\ \cmidrule{2-7}
        & \textbf{w-}$\boldsymbol{F_1}$ & \textbf{R@10} & \textbf{w-}$\boldsymbol{F_1}$ & \textbf{R@10} & \textbf{AUC} & $\boldsymbol{F_1}$ \\ \cmidrule{1-7}
        \modelnamen       & \textbf{25.74} & \textbf{40.46} & \textbf{64.18} & \textbf{80.05} & \textbf{86.04} & \textbf{74.27} \\
        \modelnamen${_b}$ & 25.46 & 39.54 & 62.75 & 78.86 & 85.20 & 72.28 \\
        \modelnamen${_c}$ & 25.21 & 39.21 & 62.52 & 78.51 & 84.47 & 71.58 \\
        \modelnamen${_d}$ & 23.92 & 37.91 & 58.75 & 77.58 & 83.18 & 71.36 \\
        \modelnamen${_e}$ & 22.12 & 36.74 & 58.45 & 77.21 & 82.35 & 70.67 \\
        \modelnamen${_f}$ & 25.37 & 39.28 & 63.01 & 78.62 & 84.51 & 73.26 \\
        {\modelnamen${_g}$} & 23.18 & 37.05 & 58.26 & 77.39 & 82.80 & 71.02 \\
        \bottomrule
    \end{tabular}
\end{table}

We report the results of {\modelnamen} and {\modelnamen}$_{b\sim g}$ on two tasks in Table~\ref{tab:ablation}. We can firstly observe that the complete {\modelnamen} achieves the best performance among {\modelnamen}$_{b\sim g}$, which proves the effectiveness of our proposed model. The comparison between {\modelnamen} and {\modelnamen}$_{b}$, and {\modelnamen}$_{d}$ and {\modelnamen}$_{e}$ shows that the hyperbolic embedding can effectively initialize the representations of medical codes. The results of {\modelnamen}$_{b}$ and {\modelnamen}$_{c}$ indicates that the hierarchical structure of medical codes is helpful to guide the prediction. We can also see that the performance of {\modelnamen}$_{c}$ has a large improvement over {\modelnamen}$_{e}$, which shows the importance of self-supervised learning. However, the performance improvement of {\modelnamen}$_{c}$ over {\modelnamen}$_{d}$ on HF prediction is not as significant as the improvement over {\modelnamen}$_{e}$. We infer that the hyperbolic embedding is also effective without self-supervised learning when the number of predicted medical codes is small. In addition, after replacing the multi-level attention with T-LSTM, the performance of {\modelnamen}$_{f}$ is not as good as {\modelnamen}. {Moreover, the result of {\modelnamen}$_{g}$ shows that the proposed {\modelnamen} graph and GNN are also important in predicting health events, because the graph structure was utilized to capture  hidden relations of diseases.} Finally, even without self-supervised learning and hyperbolic embedding ({\modelnamen}$_{e}$), our model still achieves top performance among baselines as shown in Table~\ref{tab:result_code} and~\ref{tab:result_hf}, which demonstrates the effectiveness of the proposed framework.

\begin{figure}
    \subfloat[w-$F_1$]{
        \includegraphics[width=0.47\linewidth]{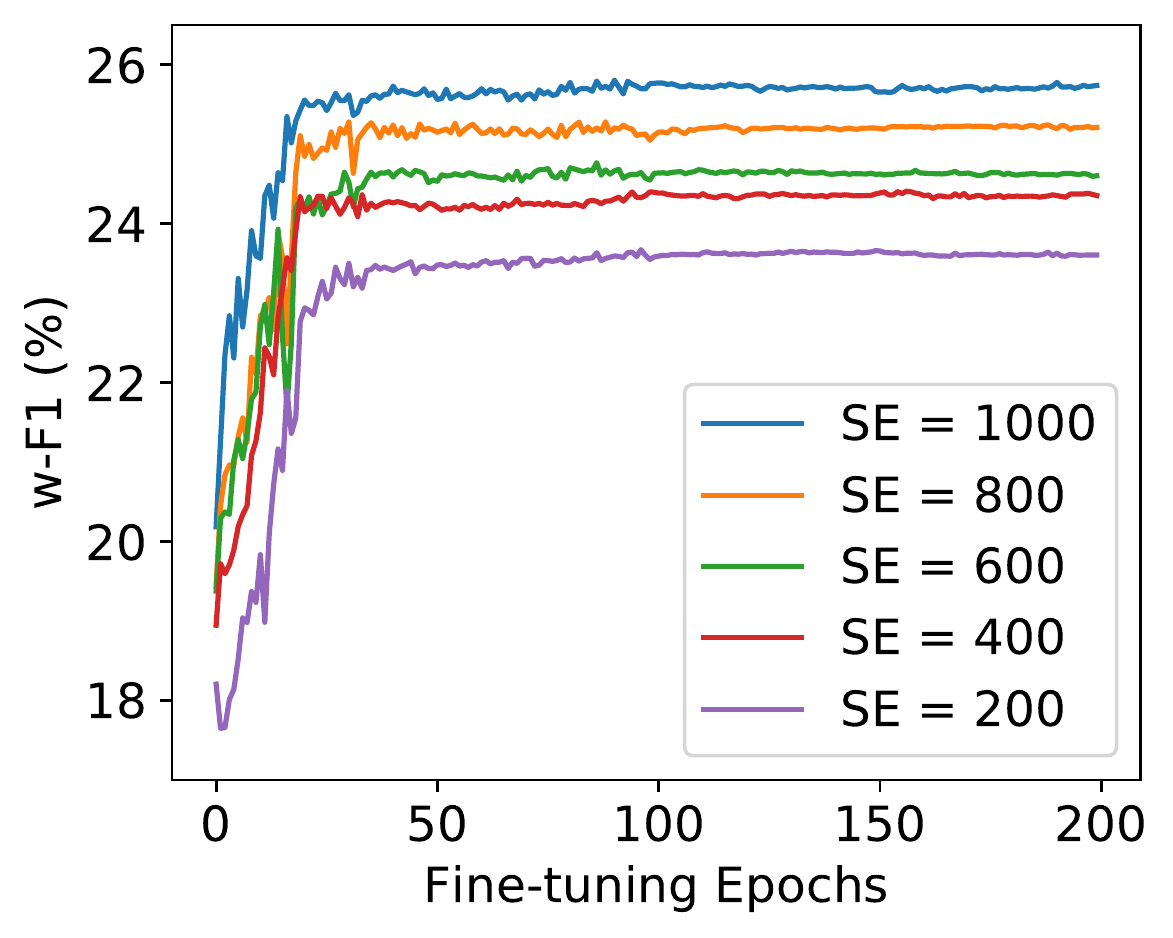}
        \label{fig:wf1}
    } \hfill
    \subfloat[R@10]{
        \includegraphics[width=0.47\linewidth]{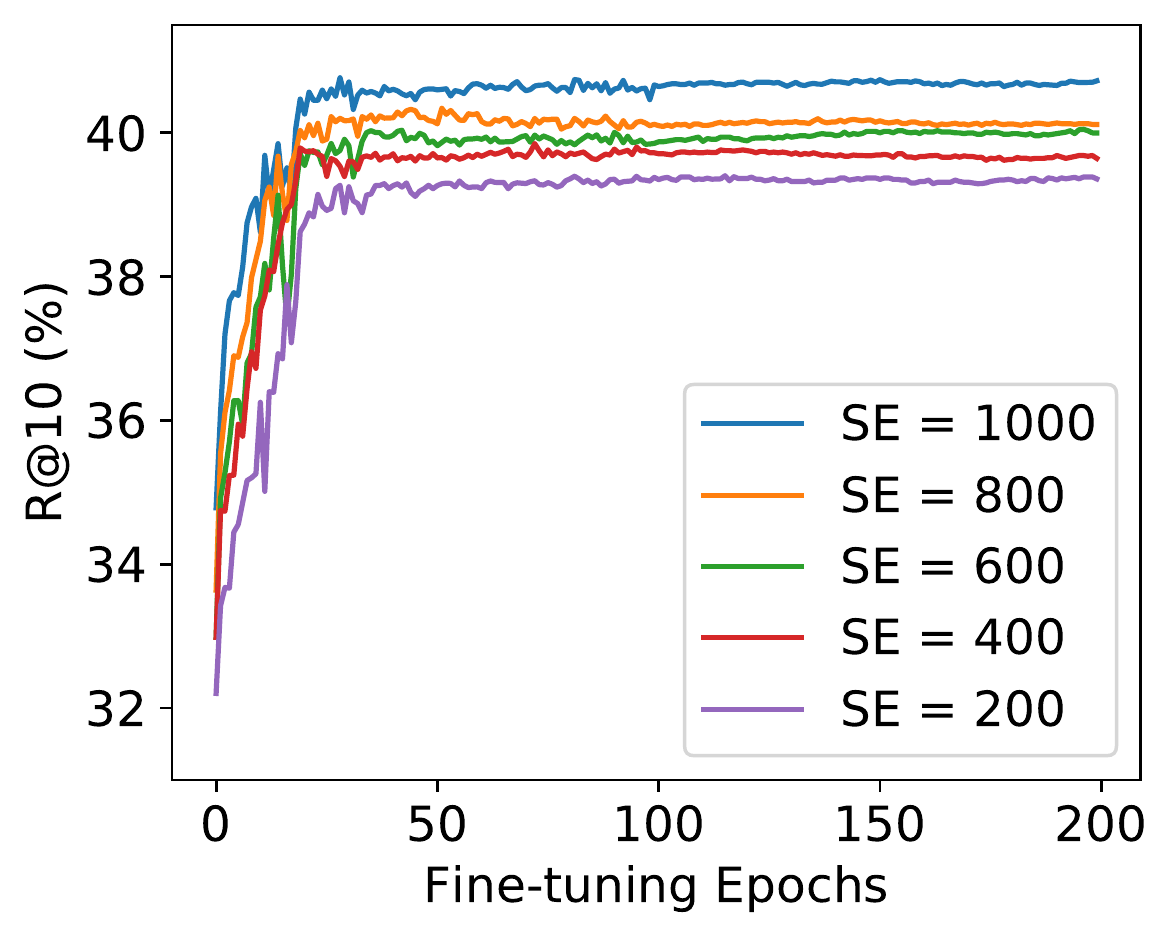}
        \label{fig:r@10}
    } \\
    \subfloat[AUC]{
        \includegraphics[width=0.47\linewidth]{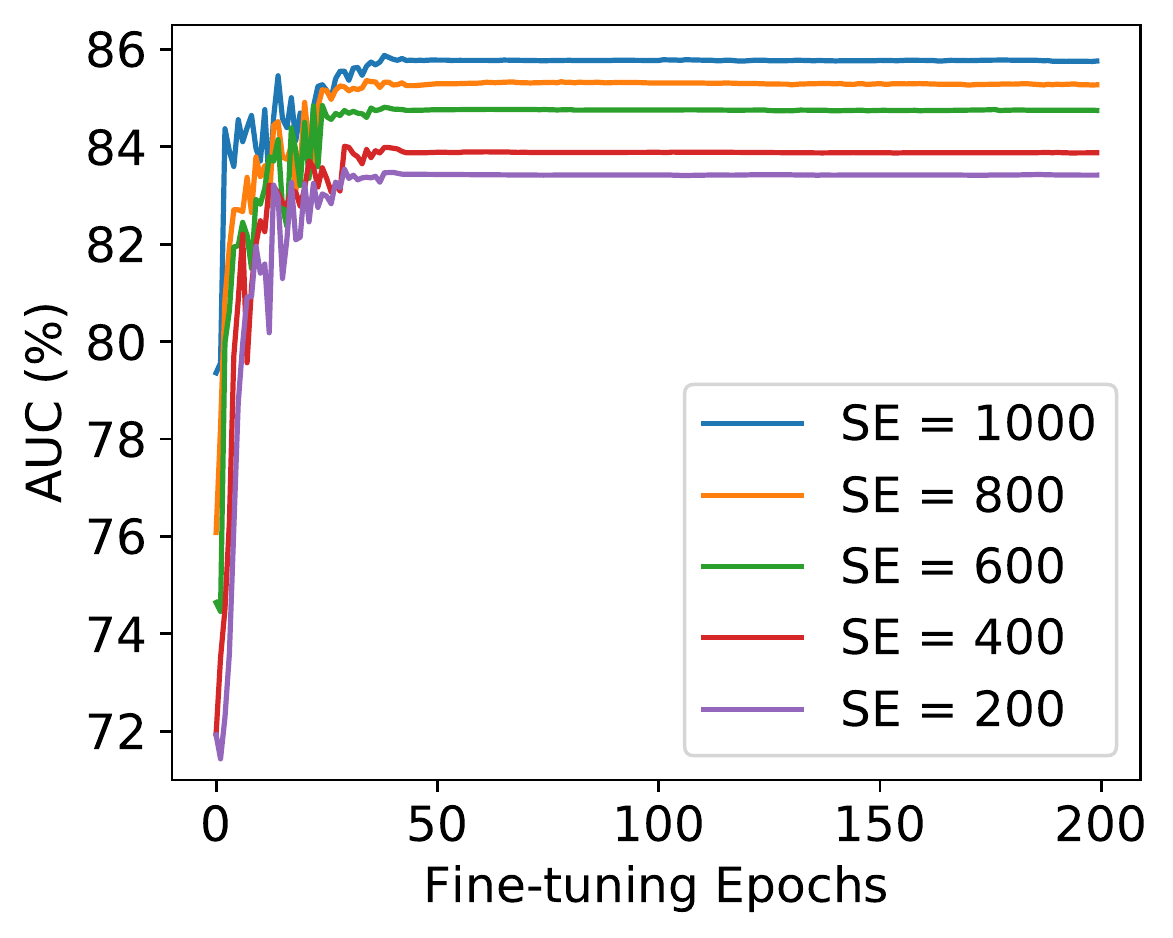}
        \label{fig:auc}
    } \hfill
    \subfloat[$F_1$]{
        \includegraphics[width=0.47\linewidth]{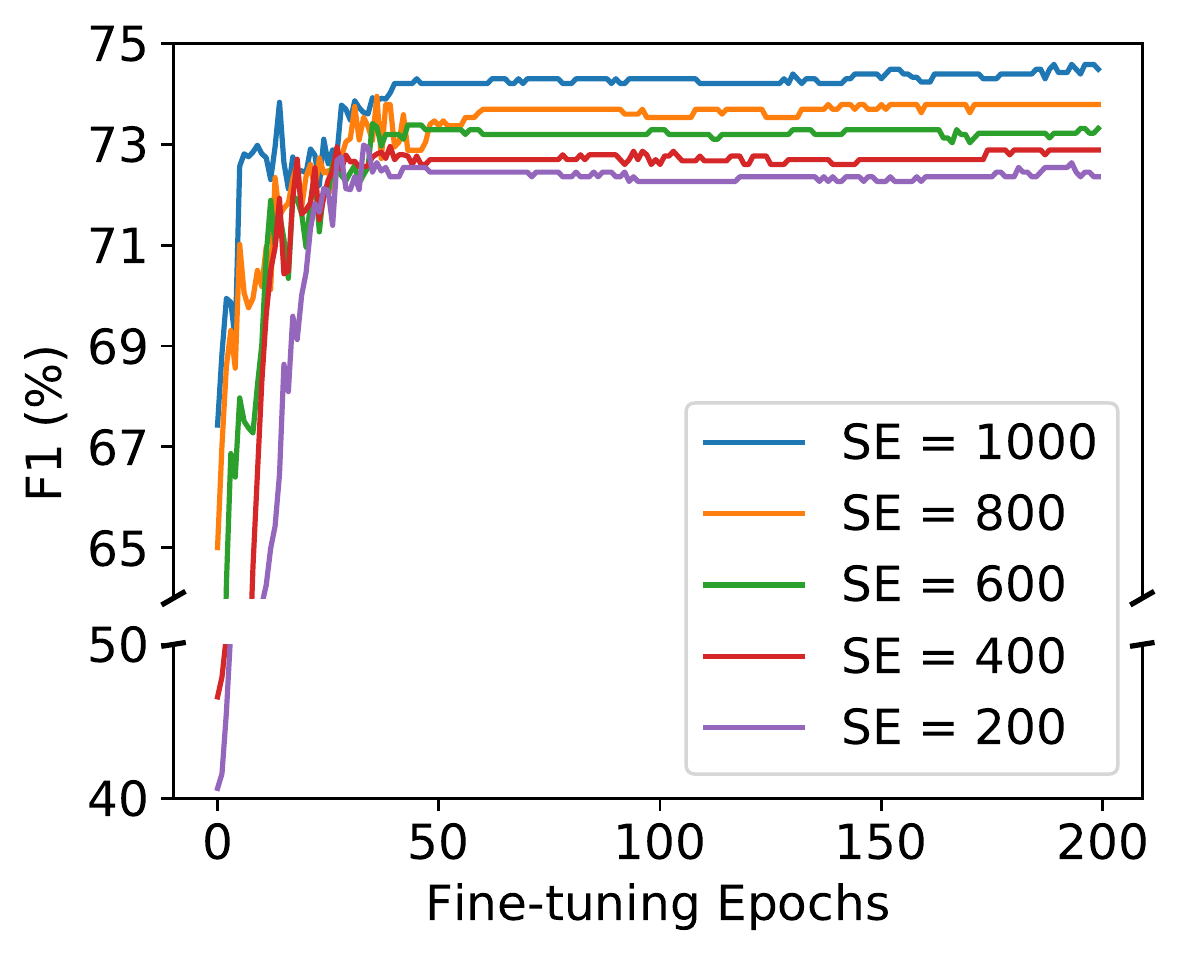}
        \label{fig:f1}
    }
    \caption{The prediction results {for fine-tuning} using w-$F_1$, R@10 on the diagnosis prediction, and AUC, $F_1$ on the HF prediction with five different self-supervised learning epochs (SE).}
    \label{fig:sl_study}
\end{figure}

\subsection{Self-supervised Learning Study}
To further study the effectiveness and impact of self-supervised learning, we adopt different self-supervised learning epochs (SE). On both tasks, we select five different SE as [200, 400, 600, 800, 1000] to report the variation of w-$F_1$, R@10 of the diagnosis prediction, and AUC and $F_1$ of the HF prediction on the validation data of both tasks in MIMIC-III during the fine-tuning phase.

\figurename{~\ref{fig:sl_study}} shows the validation results {for fine-tuning} of the diagnosis prediction and the HF prediction on different self-supervised learning epochs. {The x-axis of each subplot denotes the fine-tuning epochs, and the y-axis means each evaluation metrics for fine-tuning.} In \figurename{~\subref*{fig:wf1}} and~\subref*{fig:r@10} of the diagnosis prediction results, we can see that SE has a significant influence on the fine-tuning results. Larger SE not only produces higher w-$F_1$ and R@10, but also speeds up the convergence. As shown in \figurename{~\subref*{fig:auc}} and~\subref*{fig:f1} of the HF prediction, the final values of AUC and $F_1$ are not affected too much by SE compared to the diagnosis prediction. However, lower self-supervised learning epochs still gives worse results. Therefore, we can further conclude that self-supervised learning on the historical hierarchy prediction task helps in improving the prediction performance on different tasks and accelerates the {fine-tuning} of the model.

\subsection{Case Studies for Model Interpretation}
\label{sec:exp_interpret}
We interpret {\modelnamen} by demonstrating the representation of some typical disease complications learned by {\modelnamen} on MIMIC-III, and visualizing the contributions introduced in Section~\ref{sec:method_interpret}.

\subsubsection{Representation of disease complication}
\begin{figure}
    \subfloat[Heart related conditions]{
        \includegraphics[width=0.47\linewidth]{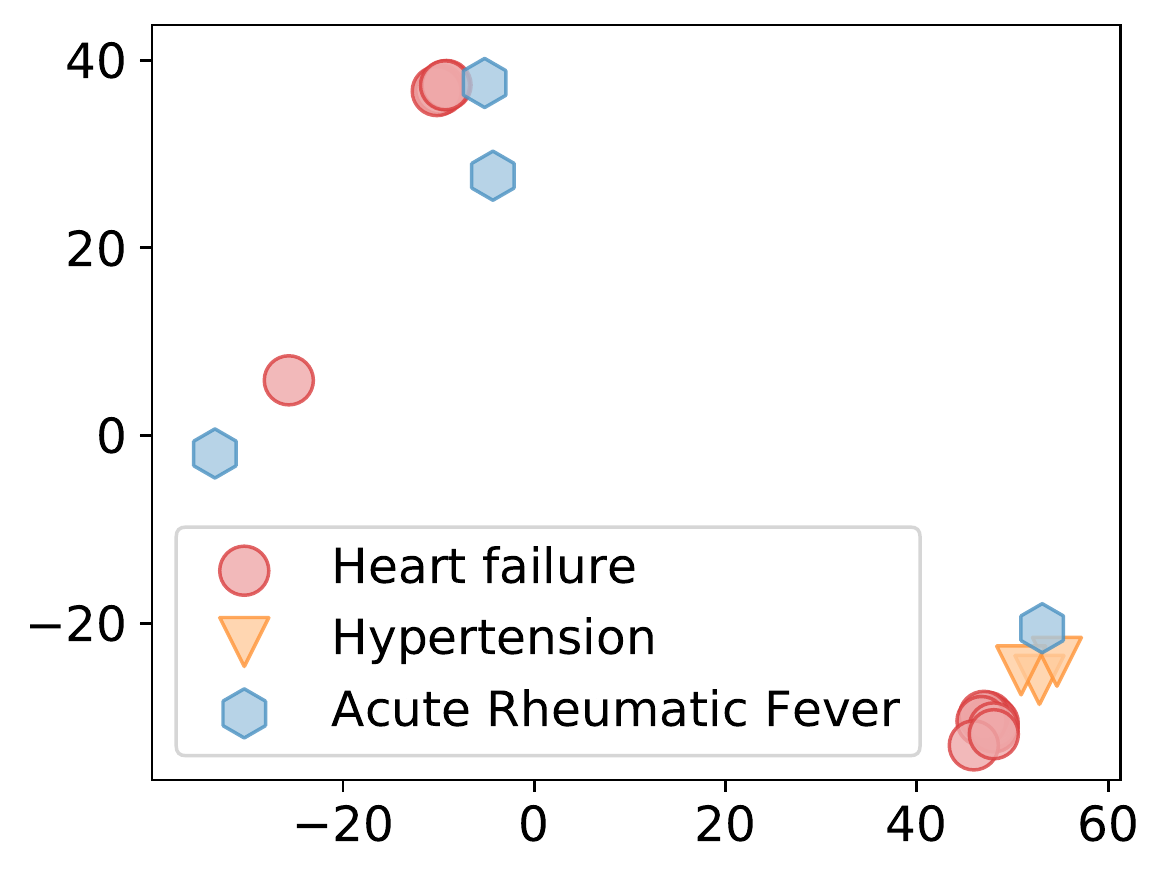}
        \label{fig:hf_hp}
    } \hfill
    \subfloat[Diabetes]{
        \includegraphics[width=0.47\linewidth]{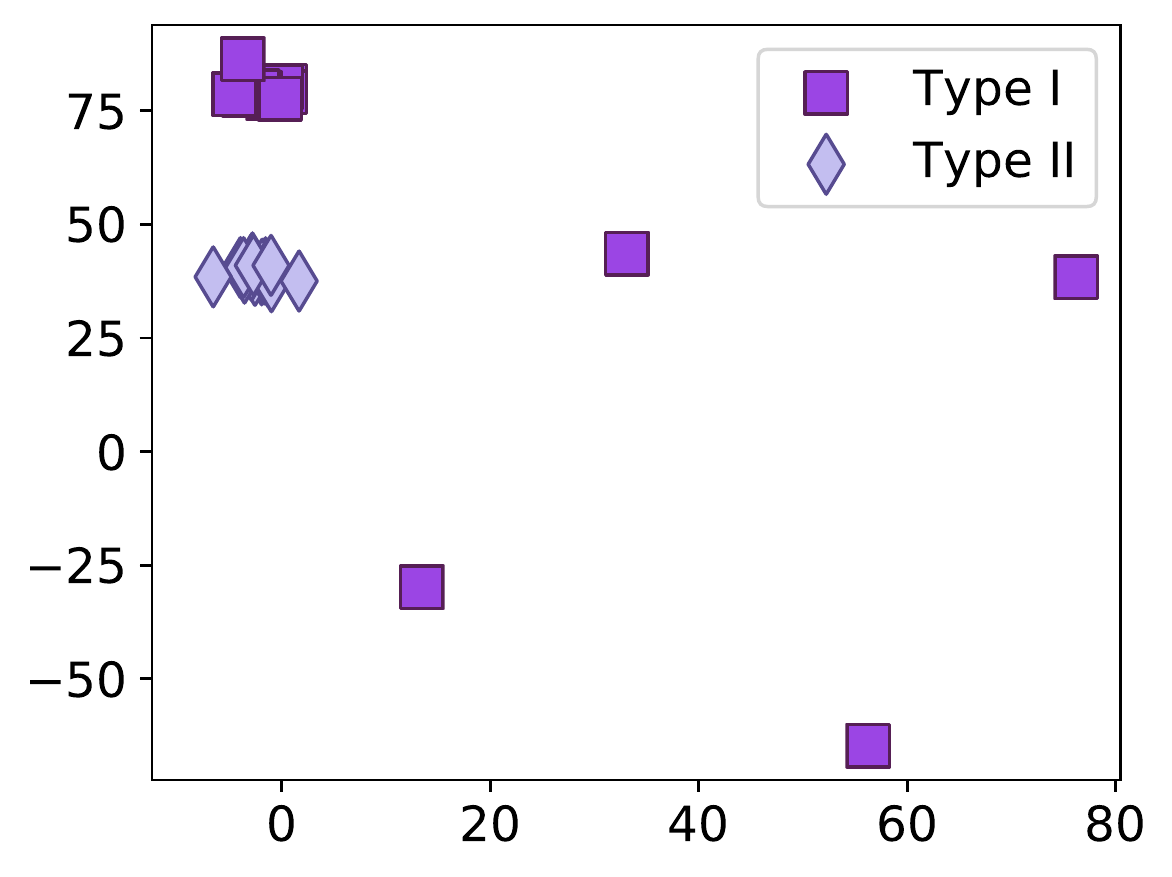}
        \label{fig:dia}
    }
    \caption{Scatter plot of heart related conditions and different diabetes using t-SNE. (a) The complications between different types of heart failure, hypertension, and acute rheumatic fever (which is a disease that can affect the heart). (b) Diabetes type I and type II.}
    \label{fig:diseases}
\end{figure}

In order to demonstrate if {\modelnamen} successfully captures the disease complications, we adopt the embedding matrix $\mathbf{X}$, i.e., the output of the graph neural network, as the final representation learned by self-supervised learning. Then, we use t-SNE~\cite{maaten2008visualizing} to project $\mathbf{X}$ into two dimensions. In the next step, we select 15 types of heart failure\footnote{\url{www.icd9data.com/2015/Volume1/390-459/420-429/428/}}, 3 types of essential hypertension\footnote{\url{www.icd9data.com/2015/Volume1/390-459/401-405/401/}}, and 7 types of acute rheumatic fever\footnote{\url{www.icd9data.com/2015/Volume1/390-459/390-392/}} that appear in MIMIC-III and plot them in \figurename{~\subref*{fig:hf_hp}}. As shown in \figurename{~\subref*{fig:hf_hp}}, they are mainly grouped into three clusters. In addition, hypertension locates near one cluster of heart failure, and acute rheumatic fever locates near the three clusters of heart failure. Given that hypertension and acute rheumatic fever are two common comorbidities of heart-related diseases, they are often diagnosed in the same set of patients. It indicates that the proposed method learns similar embeddings for similar diseases or complications. Next, we select two types of diabetes\footnote{\url{www.icd9data.com/2015/Volume1/240-279/249-259/250/}}: type I and type II, and plot their embeddings in \figurename{~\subref*{fig:dia}}. As shown in \figurename{~\subref*{fig:dia}}, there are no joint clusters between diabetes type I and diabetes type II. We infer that the complications of diabetes type I and diabetes type II are different to some extent. To summarize, {\modelnamen} is able to capture the disease complications. It can also distinguish different types of diseases based on their different complications. Therefore, {\modelnamen} can provide the general interpretability for diseases using the learned hidden representations of medical codes.

\subsubsection{Quantification of multi-level attention}
\begin{figure}
    \includegraphics[width=\linewidth]{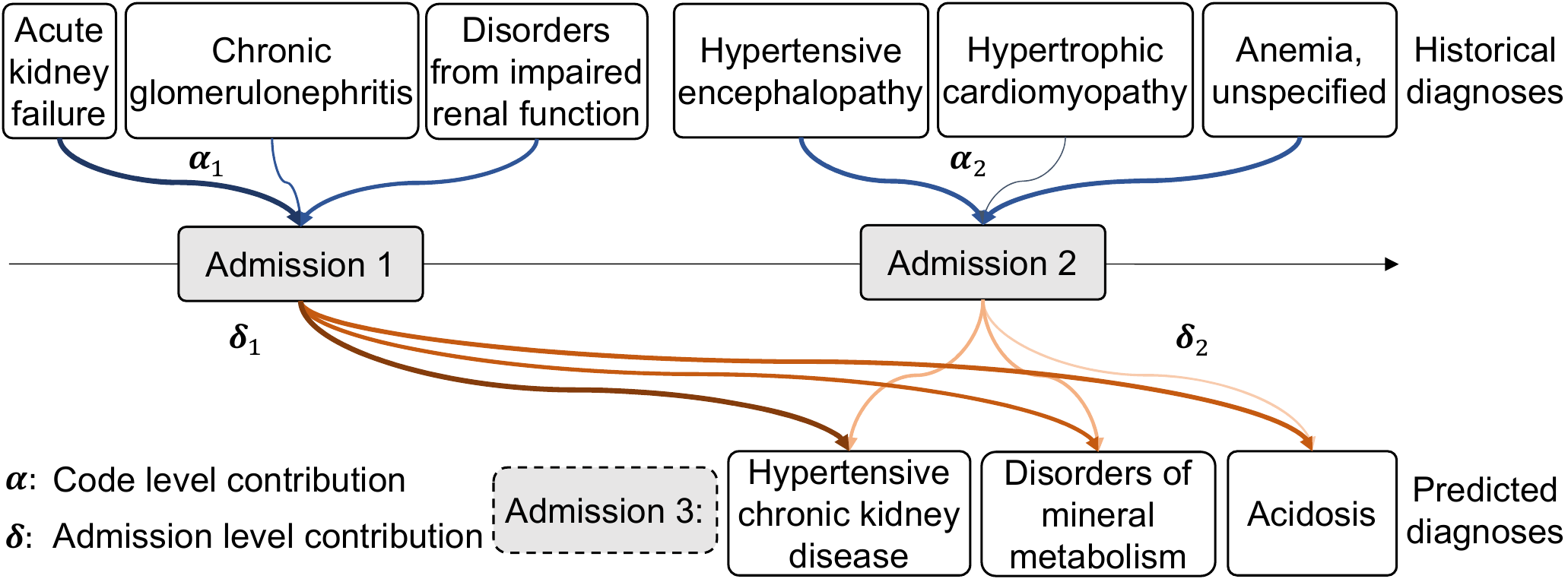}
    \caption{Contribution of diagnoses to each admission, and admissions to each predicted diagnosis. In this case, the patient has two historical admissions. We use two admissions to predict diagnoses in the third admission.}
    \label{fig:attention_weight}
\end{figure}

To measure the contributions both at the code-level and at the admission-level explained in Section~\ref{sec:method_interpret}, we visualize the code-level attention distribution $\boldsymbol{\alpha}$ in code-level attention, and coefficient $\boldsymbol{\delta}$ in Equation~(\ref{eq:coeff}) for a given patient $u$ on the diagnosis prediction task. \figurename{~\ref{fig:attention_weight}} demonstrates the historical diagnoses, the admissions, and the predicted diagnoses for a patient in rectangles. The contribution $\boldsymbol{\alpha}$ and $\boldsymbol{\delta}$ are represented by blue and red lines, respectively. The thickness and darkness of the line denote different values of $\boldsymbol{\alpha}$ and $\boldsymbol{\delta}$. Thicker and darker lines correspond to larger values of $\boldsymbol{\alpha}$ and $\boldsymbol{\delta}$. In this case, there are 6 diagnoses in the first admission and 10 diagnoses in the second admission, and we select 3 important diagnoses for both admissions. There are also 6 diagnoses of the third admission of this patient, i.e., ground truth diagnoses, and we show 3 correct predictions in \figurename{~\ref{fig:attention_weight}}.

In the first admission, the three diagnoses are all kidney-related diseases. Acute kidney failure and disorders resulting from impaired renal function mainly contribute to this admission. In the second admission, the three diagnoses belong to brain, heart, and blood-related diseases, respectively. Hypertensive encephalopathy and Anemia mainly contribute to this admission. In the third admission, we can see that the correct predictions are all metabolic diseases and kidney disease. Therefore, {\modelnamen} predicts that the first admission contributes more than the second admission. More specifically, the first admission contributes mainly to hypertensive chronic kidney disease because the patient is diagnosed with acute kidney failure in the first admission. We can also infer that kidney problems of this patient mainly cause the metabolic diseases in the third admission. For the second admission, we can discover that it mainly contributes to hypertensive chronic kidney disease. We believe that it is due to the fact this patient has hypertensive encephalopathy. High blood pressure causing hypertensive encephalopathy also causes hypertensive chronic kidney disease in this admission. Besides, anemia in the second admission can also be related to the metabolic diseases in the third admission. In summary, {\modelnamen} provides quantitative and personalized interpretability using the contributions $\boldsymbol{\alpha}$ and $\boldsymbol{\delta}$ learned from patients' historical diagnoses and admissions.

\section{Conclusion}\label{sec:conclusion}
In this paper, we propose \modelnamen, a \textit{self-supervised graph learning framework with hyperbolic embeddings for medical codes to predict temporal health events}. We first take advantage of the hierarchical structure of medical codes to pre-train a hyperbolic embedding for diseases. Then, we adopt a graph neural network with a weighted and directed graph of medical codes to learn disease complications in EHR data. With the specially-designed code-level and admission-level attention mechanism, \modelnamen~is able to simultaneously provide generic interpretability for medical concepts and personalized interpretability for patients. In addition, we also design a self-supervised proxy task to predict the historical hierarchy of diagnoses in patients' admission records by further utilizing the hierarchical structure of medical codes. This task is able to  leverage more data by incorporating single admission records and the final admissions of multiple admission records. Our experimental results show the improved performance of \modelnamen~over state-of-the-art methods. We also illustrate the generic and personalized interpretability of \modelnamen~using case studies. One shortcoming of \modelnamen~is that it only utilizes disease codes in patients' admission records. In the future, we will explore the relationships of more features in EHR data, such as procedures, medicines, and lab results.

\section*{Acknowledgment}
This work is supported in part by the US National Science Foundation under grants 1948432 and 2047843. Any opinions, findings, and conclusions or recommendations expressed in this material are those of the authors and do not necessarily reflect the views of the National Science Foundation.

\ifCLASSOPTIONcaptionsoff
  \newpage
\fi

\bibliographystyle{IEEEtran}
\bibliography{ref}

\begin{thebibliography}{10}
\providecommand{\url}[1]{#1}
\csname url@samestyle\endcsname
\providecommand{\newblock}{\relax}
\providecommand{\bibinfo}[2]{#2}
\providecommand{\BIBentrySTDinterwordspacing}{\spaceskip=0pt\relax}
\providecommand{\BIBentryALTinterwordstretchfactor}{4}
\providecommand{\BIBentryALTinterwordspacing}{\spaceskip=\fontdimen2\font plus
\BIBentryALTinterwordstretchfactor\fontdimen3\font minus
  \fontdimen4\font\relax}
\providecommand{\BIBforeignlanguage}[2]{{%
\expandafter\ifx\csname l@#1\endcsname\relax
\typeout{** WARNING: IEEEtran.bst: No hyphenation pattern has been}%
\typeout{** loaded for the language `#1'. Using the pattern for}%
\typeout{** the default language instead.}%
\else
\language=\csname l@#1\endcsname
\fi
#2}}
\providecommand{\BIBdecl}{\relax}
\BIBdecl

\bibitem{choi2016doctor}
E.~Choi, M.~T. Bahadori, A.~Schuetz, W.~F. Stewart, and J.~Sun, ``Doctor ai:
  Predicting clinical events via recurrent neural networks,'' in \emph{Machine
  Learning for Healthcare Conference}, 2016, pp. 301--318.

\bibitem{zheng2020predicting}
N.~Zheng, S.~Du, J.~Wang, H.~Zhang, W.~Cui, Z.~Kang, T.~Yang, B.~Lou, Y.~Chi,
  H.~Long \emph{et~al.}, ``Predicting covid-19 in china using hybrid ai
  model,'' \emph{IEEE Transactions on Cybernetics}, 2020.

\bibitem{choi2017gram}
E.~Choi, M.~T. Bahadori, L.~Song, W.~F. Stewart, and J.~Sun, ``Gram:
  graph-based attention model for healthcare representation learning,'' in
  \emph{Proceedings of the 23rd ACM SIGKDD International Conference on
  Knowledge Discovery \& Data Mining}, 2017, pp. 787--795.

\bibitem{bai2018interpretable}
T.~Bai, S.~Zhang, B.~L. Egleston, and S.~Vucetic, ``Interpretable
  representation learning for healthcare via capturing disease progression
  through time,'' in \emph{Proceedings of the 24th ACM SIGKDD International
  Conference on Knowledge Discovery \& Data Mining}, 2018, pp. 43--51.

\bibitem{xu2021comorbidity}
Z.~Xu, J.~Zhang, Q.~Zhang, Q.~Xuan, and P.~S.~F. Yip, ``A comorbidity
  knowledge-aware model for disease prognostic prediction,'' \emph{IEEE
  Transactions on Cybernetics}, 2021.

\bibitem{nguyen2018effective}
D.~Nguyen, W.~Luo, S.~Venkatesh, and D.~Phung, ``Effective identification of
  similar patients through sequential matching over icd code embedding,''
  \emph{Journal of medical systems}, vol.~42, no.~5, p.~94, 2018.

\bibitem{yang2020cross}
B.~Yang, M.~Ye, Q.~Tan, and P.~C. Yuen, ``Cross-domain missingness-aware
  time-series adaptation with similarity distillation in medical
  applications,'' \emph{IEEE Transactions on Cybernetics}, 2020.

\bibitem{darabi2019taper}
S.~Darabi, M.~Kachuee, S.~Fazeli, and M.~Sarrafzadeh, ``Taper: Time-aware
  patient ehr representation,'' \emph{IEEE Journal of Biomedical and Health
  Informatics}, vol.~24, no.~11, pp. 3268--3275, 2020.

\bibitem{mulyadi2021uncertainty}
A.~W. Mulyadi, E.~Jun, and H.-I. Suk, ``Uncertainty-aware variational-recurrent
  imputation network for clinical time series,'' \emph{IEEE Transactions on
  Cybernetics}, 2021.

\bibitem{Phuoc2017deepr}
P.~{Nguyen}, T.~{Tran}, N.~{Wickramasinghe}, and S.~{Venkatesh}, ``$\mathtt
  {Deepr}$: A convolutional net for medical records,'' \emph{IEEE Journal of
  Biomedical and Health Informatics}, vol.~21, no.~1, pp. 22--30, Jan 2017.

\bibitem{huang2017regularized}
Z.~Huang, W.~Dong, H.~Duan, and J.~Liu, ``A regularized deep learning approach
  for clinical risk prediction of acute coronary syndrome using electronic
  health records,'' \emph{IEEE Transactions on Biomedical Engineering},
  vol.~65, no.~5, pp. 956--968, 2017.

\bibitem{che2017boosting}
Z.~Che, Y.~Cheng, S.~Zhai, Z.~Sun, and Y.~Liu, ``Boosting deep learning risk
  prediction with generative adversarial networks for electronic health
  records,'' in \emph{2017 IEEE International Conference on Data Mining
  (ICDM)}.\hskip 1em plus 0.5em minus 0.4em\relax IEEE, 2017, pp. 787--792.

\bibitem{wang2019data}
C.~Wang, N.~Lu, Y.~Cheng, and B.~Jiang, ``A data-driven aero-engine degradation
  prognostic strategy,'' \emph{IEEE Transactions on Cybernetics}, vol.~51,
  no.~3, pp. 1531--1541, 2021.

\bibitem{yao2018topic}
L.~Yao, Y.~Zhang, B.~Wei, W.~Zhang, and Z.~Jin, ``A topic modeling approach for
  traditional chinese medicine prescriptions,'' \emph{IEEE Transactions on
  Knowledge and Data Engineering}, vol.~30, no.~6, pp. 1007--1021, 2018.

\bibitem{shang2019pretrain}
J.~Shang, T.~Ma, C.~Xiao, and J.~Sun, ``Pre-training of graph augmented
  transformers for medication recommendation,'' in \emph{Proceedings of the
  Twenty-Eighth International Joint Conference on Artificial Intelligence,
  {IJCAI-19}}.\hskip 1em plus 0.5em minus 0.4em\relax ijcai.org, 2019, pp.
  5953--5959.

\bibitem{johnson2016mimic}
A.~E. Johnson, T.~J. Pollard, L.~Shen, H.~L. Li-wei, M.~Feng, M.~Ghassemi,
  B.~Moody, P.~Szolovits, L.~A. Celi, and R.~G. Mark, ``Mimic-iii, a freely
  accessible critical care database,'' \emph{Scientific data}, vol.~3, p.
  160035, 2016.

\bibitem{icd9cm}
\BIBentryALTinterwordspacing
CDC, ``Icd-9-cm - international classification of diseases, ninth revision,
  clinical modification,'' Nov 2015, accessed: 2020-05-10. [Online]. Available:
  \url{https://www.cdc.gov/nchs/icd/icd9cm.htm}
\BIBentrySTDinterwordspacing

\bibitem{messerli2017transition}
F.~H. Messerli, S.~F. Rimoldi, and S.~Bangalore, ``The transition from
  hypertension to heart failure: contemporary update,'' \emph{JACC: Heart
  Failure}, vol.~5, no.~8, pp. 543--551, 2017.

\bibitem{choi2016retain}
E.~Choi, M.~T. Bahadori, J.~Sun, J.~Kulas, A.~Schuetz, and W.~Stewart,
  ``Retain: An interpretable predictive model for healthcare using reverse time
  attention mechanism,'' in \emph{Advances in Neural Information Processing
  Systems}, 2016, pp. 3504--3512.

\bibitem{ma2017dipole}
F.~Ma, R.~Chitta, J.~Zhou, Q.~You, T.~Sun, and J.~Gao, ``Dipole: Diagnosis
  prediction in healthcare via attention-based bidirectional recurrent neural
  networks,'' in \emph{Proceedings of the 23rd ACM SIGKDD
  international~conference on knowledge discovery \& data mining}, 2017, pp.
  1903--1911.

\bibitem{cho2014learning}
K.~Cho, B.~van Merri{\"e}nboer, C.~Gulcehre, D.~Bahdanau, F.~Bougares,
  H.~Schwenk, and Y.~Bengio, ``Learning phrase representations using rnn
  encoder-decoder for statistical machine translation,'' in \emph{Proceedings
  of the 2014 conference on empirical methods in natural language processing
  (EMNLP)}, 2014, pp. 1724--1734.

\bibitem{luo2020hitanet}
J.~Luo, M.~Ye, C.~Xiao, and F.~Ma, ``Hitanet: Hierarchical time-aware attention
  networks for risk prediction on electronic health records,'' in
  \emph{Proceedings of the 26th ACM SIGKDD International Conference on
  Knowledge Discovery \& Data Mining}, 2020, pp. 647--656.

\bibitem{jing2020self}
L.~Jing and Y.~Tian, ``Self-supervised visual feature learning with deep neural
  networks: A survey,'' \emph{IEEE Transactions on Pattern Analysis and Machine
  Intelligence}, 2020.

\bibitem{gidaris2018unsupervised}
\BIBentryALTinterwordspacing
S.~Gidaris, P.~Singh, and N.~Komodakis, ``Unsupervised representation learning
  by predicting image rotations,'' in \emph{International Conference on
  Learning Representations}, 2018. [Online]. Available:
  \url{https://openreview.net/forum?id=S1v4N2l0-}
\BIBentrySTDinterwordspacing

\bibitem{xu2019ternary}
X.~Xu, H.~Lu, J.~Song, Y.~Yang, H.~T. Shen, and X.~Li, ``Ternary adversarial
  networks with self-supervision for zero-shot cross-modal retrieval,''
  \emph{IEEE transactions on cybernetics}, vol.~50, no.~6, pp. 2400--2413,
  2019.

\bibitem{vaswani2017attention}
A.~Vaswani, N.~Shazeer, N.~Parmar, J.~Uszkoreit, L.~Jones, A.~N. Gomez,
  {\L}.~Kaiser, and I.~Polosukhin, ``Attention is all you need,'' in
  \emph{Advances in neural information processing systems}, 2017, pp.
  5998--6008.

\bibitem{devlin2018bert}
J.~Devlin, M.-W. Chang, K.~Lee, and K.~Toutanova, ``{BERT}: Pre-training of
  deep bidirectional transformers for language understanding,'' in
  \emph{Proceedings of the 2019 Conference of the North {A}merican Chapter of
  the Association for Computational Linguistics: Human Language Technologies,
  Volume 1 (Long and Short Papers)}.\hskip 1em plus 0.5em minus 0.4em\relax
  Association for Computational Linguistics, Jun. 2019, pp. 4171--4186.

\bibitem{pennington2014glove}
J.~Pennington, R.~Socher, and C.~D. Manning, ``Glove: Global vectors for word
  representation,'' in \emph{Proceedings of the 2014 conference on empirical
  methods in natural language processing (EMNLP)}, 2014, pp. 1532--1543.

\bibitem{yu2012adaptive}
J.~Yu, D.~Tao, and M.~Wang, ``Adaptive hypergraph learning and its application
  in image classification,'' \emph{IEEE Transactions on Image Processing},
  vol.~21, no.~7, pp. 3262--3272, 2012.

\bibitem{kipf2016semi}
T.~N. Kipf and M.~Welling, ``{Semi-Supervised Classification with Graph
  Convolutional Networks},'' in \emph{Proceedings of the 5th International
  Conference on Learning Representations}, 2017.

\bibitem{nickel2017poincare}
M.~Nickel and D.~Kiela, ``Poincar{\'e} embeddings for learning hierarchical
  representations,'' in \emph{Advances in neural information processing
  systems}, 2017, pp. 6338--6347.

\bibitem{chami2020low}
I.~Chami, A.~Wolf, D.-C. Juan, F.~Sala, S.~Ravi, and C.~R{\'e},
  ``Low-dimensional hyperbolic knowledge graph embeddings,'' \emph{arXiv
  preprint arXiv:2005.00545}, 2020.

\bibitem{choudhary2021self}
N.~Choudhary, N.~Rao, S.~Katariya, K.~Subbian, and C.~K. Reddy,
  ``Self-supervised hyperboloid representations from logical queries over
  knowledge graphs,'' in \emph{Proceedings of the Web Conference}, 2021, pp.
  1373--1384.

\bibitem{gct_aaai20}
E.~Choi, Z.~Xu, Y.~Li, M.~W. Dusenberry, G.~Flores, Y.~Xue, and A.~M. Dai,
  ``Learning the graphical structure of electronic health records with graph
  convolutional transformer,'' in \emph{Proceedings of the 34th Conference on
  Association for the Advancement of Artificial Intelligence}, 2020.

\bibitem{lu2021collaborative}
C.~Lu, C.~K. Reddy, P.~Chakraborty, S.~Kleinberg, and Y.~Ning, ``Collaborative
  graph learning with auxiliary text for temporal event prediction in
  healthcare,'' in \emph{Proceedings of the Thirtieth International Joint
  Conference on Artificial Intelligence, {IJCAI-21}}, 2021.

\bibitem{icd10}
\BIBentryALTinterwordspacing
CDC, ``Icd-10 - international classification of diseases, tenth revision,'' Feb
  2020, accessed: 2020-05-10. [Online]. Available:
  \url{https://www.cdc.gov/nchs/icd/icd10cm.htm}
\BIBentrySTDinterwordspacing

\bibitem{dhingra2018embedding}
B.~Dhingra, C.~Shallue, M.~Norouzi, A.~Dai, and G.~Dahl, ``Embedding text in
  hyperbolic spaces,'' in \emph{Proceedings of the Twelfth Workshop on
  Graph-Based Methods for Natural Language Processing (TextGraphs-12)}, 2018,
  pp. 59--69.

\bibitem{luong-etal-2015-effective}
T.~Luong, H.~Pham, and C.~D. Manning, ``Effective approaches to attention-based
  neural machine translation,'' in \emph{Proceedings of the 2015 Conference on
  Empirical Methods in Natural Language Processing}.\hskip 1em plus 0.5em minus
  0.4em\relax Lisbon, Portugal: Association for Computational Linguistics, Sep.
  2015, pp. 1412--1421.

\bibitem{xu2015empirical}
B.~Xu, N.~Wang, T.~Chen, and M.~Li, ``Empirical evaluation of rectified
  activations in convolutional network,'' \emph{arXiv preprint
  arXiv:1505.00853}, 2015.

\bibitem{bishop2006pattern}
C.~M. Bishop, \emph{Pattern recognition and machine learning}.\hskip 1em plus
  0.5em minus 0.4em\relax Springer, 2006, pp. 359--361.

\bibitem{pollard2018eicu}
T.~J. Pollard, A.~E. Johnson, J.~D. Raffa, L.~A. Celi, R.~G. Mark, and
  O.~Badawi, ``The eicu collaborative research database, a freely available
  multi-center database for critical care research,'' \emph{Scientific data},
  vol.~5, p. 180178, 2018.

\bibitem{choi2018mime}
E.~Choi, C.~Xiao, W.~Stewart, and J.~Sun, ``Mime: Multilevel medical embedding
  of electronic health records for predictive healthcare,'' in \emph{Advances
  in neural information processing systems}, 2018, pp. 4547--4557.

\bibitem{chowdhury2019mixed}
S.~Chowdhury, C.~Zhang, P.~S. Yu, and Y.~Luo, ``Mixed pooling multi-view
  attention autoencoder for representation learning in healthcare,''
  \emph{arXiv preprint arXiv:1910.06456}, 2019.

\bibitem{srivastava2014dropout}
N.~Srivastava, G.~Hinton, A.~Krizhevsky, I.~Sutskever, and R.~Salakhutdinov,
  ``Dropout: a simple way to prevent neural networks from overfitting,''
  \emph{The journal of machine learning research}, vol.~15, no.~1, pp.
  1929--1958, 2014.

\bibitem{KingmaB14}
D.~P. Kingma and J.~Ba, ``Adam: {A} method for stochastic optimization,'' in
  \emph{3rd International Conference on Learning Representations, {ICLR} 2015,
  San Diego, CA, USA, May 7-9, 2015, Conference Track Proceedings}, Y.~Bengio
  and Y.~LeCun, Eds., 2015.

\bibitem{hinton2012neural}
G.~Hinton, N.~Srivastava, and K.~Swersky, ``Neural networks for machine
  learning lecture 6a overview of mini-batch gradient descent,'' \emph{Cited
  on}, vol.~14, no.~8, 2012.

\bibitem{baytas2017patient}
I.~M. Baytas, C.~Xiao, X.~Zhang, F.~Wang, A.~K. Jain, and J.~Zhou, ``Patient
  subtyping via time-aware lstm networks,'' in \emph{Proceedings of the 23rd
  ACM SIGKDD international conference on knowledge discovery \& data mining},
  2017, pp. 65--74.

\bibitem{maaten2008visualizing}
L.~v.~d. Maaten and G.~Hinton, ``Visualizing data using t-sne,'' \emph{Journal
  of machine learning research}, vol.~9, no. Nov, pp. 2579--2605, 2008.

\end{thebibliography}
\begin{IEEEbiography}[{\includegraphics[width=1in,clip]{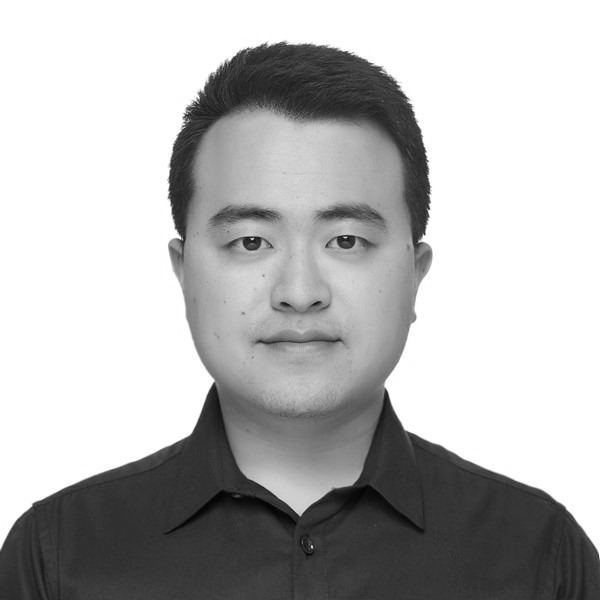}}]{Chang Lu}
is working towards his Ph.D. degree in the Department of Computer Science at Stevens Institute of Technology, advised by Dr. Yue Ning. He received his BS degree in 2016 and MS degree in 2019, in the School of Computer science, Fudan University. His research interests include big data, deep learning, and interpretable predictive models in healthcare domain.
\end{IEEEbiography}

\begin{IEEEbiography}[{\includegraphics[width=1in,clip]{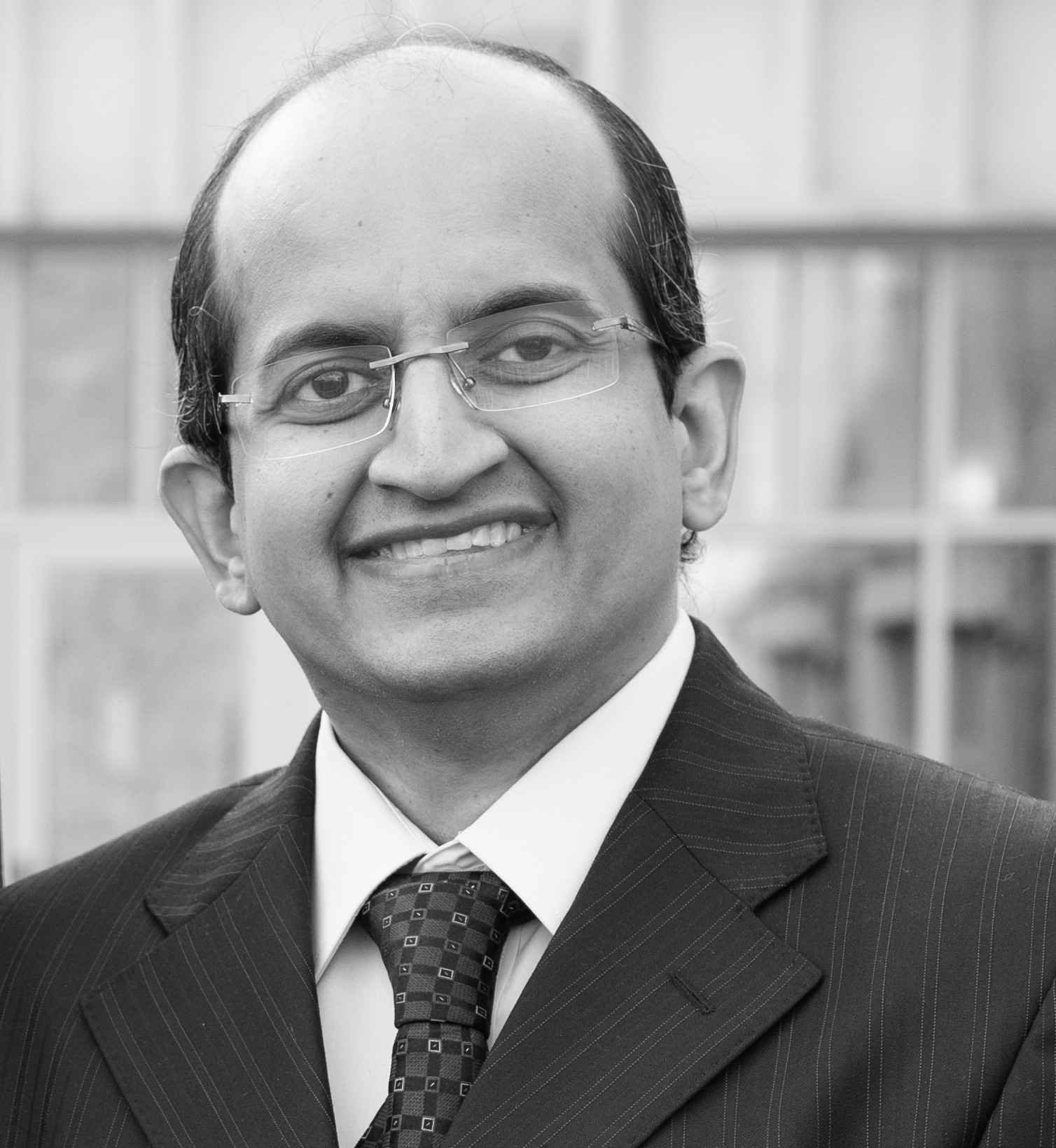}}]{Chandan K. Reddy}
is a Professor in the Department of Computer Science at Virginia Tech. He received his Ph.D. from Cornell University and M.S. from Michigan State University. His primary research interests are machine learning with applications to various real-world domains including healthcare, transportation, social networks, and e-commerce. He has published over 130 peer-reviewed articles in leading conferences and journals. He received several awards for his research work including the Best Application Paper Award at ACM SIGKDD conference in 2010, Best Poster Award at IEEE VAST conference in 2014, Best Student Paper Award at IEEE ICDM conference in 2016, and was a finalist of the INFORMS Franz Edelman Award Competition in 2011.  He is serving on the editorial boards of ACM TKDD, IEEE Big Data, and ACM TIST journals. He is a senior member of the IEEE and distinguished member of the ACM.
\end{IEEEbiography}

\begin{IEEEbiography}[{\includegraphics[width=1in,clip]{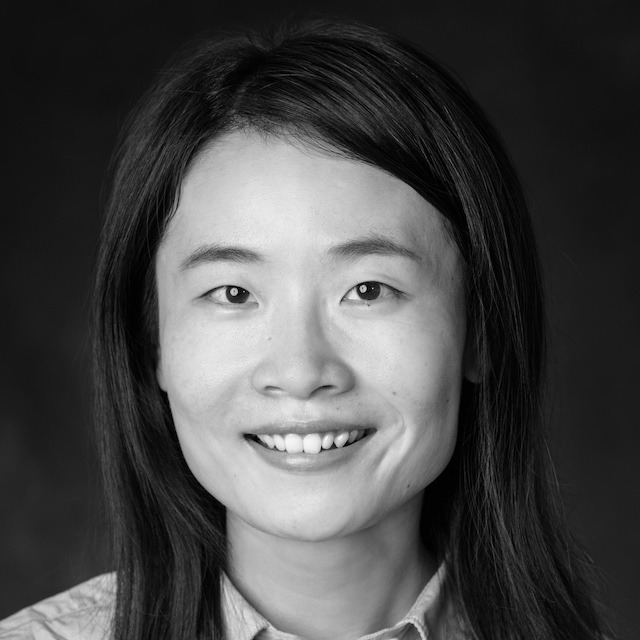}}]{Yue Ning}
is an Assistant Professor in the Department of Computer Science at Stevens Institute of Technology, where she is also affiliated with the Stevens Institute for Artificial Intelligence (SIAI). She received her PhD degree in Computer Science from Virginia Tech in 2018. Her research interests are in the general areas of machine learning, data analytics, and social media analysis. Specifically,  she is focused on developing predictive methods to capture spatiotemporal, dynamic, and interpretable patterns in large-scale data with applications in computational social science and health informatics. She has published over 20 peer-reviewed papers in prestigious journals and conferences including TKDD, KDD, AAAI, CIKM, ICWSM, and SDM. She has served on program committees of KDD, AAAI, SDM, ICML, IEEE BigData, ASONAM, and ICLR, and as a journal reviewer for TKDD, TKDE, DAMI, and PR. Her research is supported by Stevens Institute of Technology, Nvidia, and the National Science Foundation.
\end{IEEEbiography}

\end{document}